\pdfoutput=1
\documentclass[10pt,twocolumn,letterpaper]{article}

\usepackage{cvpr}              

\usepackage[dvipsnames]{xcolor}

\definecolor{cvprblue}{rgb}{0.21,0.49,0.74}
\usepackage[pagebackref,breaklinks,colorlinks,citecolor=cvprblue]{hyperref}

\usepackage{graphicx}
\usepackage{booktabs} 
\usepackage{comment}
\usepackage{amsmath,amssymb} 
\usepackage{color}
\usepackage{subcaption}
\usepackage{mathtools}
\usepackage{paralist} 
\usepackage{colortbl}
\usepackage{nicefrac}
\usepackage{multirow}
\usepackage{afterpage}
\usepackage{float}
\usepackage{bm}
\usepackage{adjustbox}

\usepackage{afterpage}
\def\paperID{*****} 
\def\confName{3DV\xspace}
\def\confYear{2025\xspace}

\title{INRet: A General Framework for Accurate Retrieval of INRs for Shapes}

\author{
Yushi Guan$^1$, Daniel Kwan$^1$, Ruofan Liang$^1$, Selvakumar Panneer$^2$,\\
Nilesh Jain$^2$, Nilesh Ahuja$^2$, Nandita Vijaykumar$^1$ \\
$^1$University of Toronto \quad $^2$Intel \\
{\tt\small \{guanyushi, dkwan, ruofan, nandita\}@cs.toronto.edu}\\
{\tt\small \{selvakumar.panneer, nilesh.jain, nilesh.ahuja\}@intel.com}
}

\newcommand{\name}{INRet\xspace}

\definecolor{cl1}{RGB}{86, 108, 161}

\definecolor{cl2}{RGB}{164, 133, 115}

\definecolor{cl3}{RGB}{98, 139, 66}
\definecolor{cl4}{RGB}{167, 129, 15}
\definecolor{cl5}{RGB}{134, 104, 154}
\definecolor{cl6}{RGB}{160, 81, 70}
\definecolor{cl7}{RGB}{200, 176, 81}

\newcommand{\lA}{\textcolor{cl1}}

\newcommand{\lB}{\textcolor{cl2}}

\newcommand{\lC}{\textcolor{cl3}}
\newcommand{\lD}{\textcolor{cl4}}
\newcommand{\lE}{\textcolor{cl5}}
\newcommand{\lF}{\textcolor{cl6}}

\begin{document}
\makeatletter\@namedef{enit@series}{}\makeatother
\maketitle
\begin{abstract}
Implicit neural representations (INRs) have become an important method for encoding various data types, such as 3D objects or scenes, images, and videos. They have proven to be particularly effective at representing 3D content, e.g., 3D scene reconstruction from 2D images, novel 3D content creation, as well as the representation, interpolation and completion of 3D shapes. With the widespread generation of 3D data in an INR format, there is a need to support effective organization and retrieval of INRs saved in a data store. A key aspect of retrieval and clustering of INRs in a data store is the formulation of similarity between INRs that would, for example, enable retrieval of similar INRs using a query INR. In this work, we propose \textbf{\name} (\textbf{INR} \textbf{Ret}rieve), a method for determining similarity between INRs that represent shapes, thus enabling accurate retrieval of similar shape INRs from an INR data store. \name flexibly supports different INR architectures such as INRs with octree grids, triplanes, and hash grids, as well as different implicit functions including signed/unsigned distance function and occupancy field. We demonstrate that our method is more general and accurate than the existing INR retrieval method, which only supports simple MLP INRs and requires the same architecture between the query and stored INRs. Furthermore, compared to converting INRs to other representations (e.g., point clouds or multi-view images) for 3D shape retrieval, \name achieves higher accuracy while avoiding the conversion overhead.

\end{abstract}
\section{Introduction}
\label{sec:intro}

Implicit neural representations (INRs) have become an important approach for representing various types of data, including images, videos, audio, and 3D content~\cite{inr_compression_coin,inr_compression_video,inraudio,nerf}. Instead of storing the data explicitly (such as RGB pixel values for images or meshes for 3D data), INRs typically encode the data implicitly by training a neural network that maps an input location to an output value. Compared to traditional representations, INRs offer several key advantages including a compact and differentiable representation, and the ability to be decoded at any resolution~\cite{imnet,deepsdf}. INRs have seen many applications including neural compression~\cite{inr_compression_coin,inr_compression_video}, super-resolution, and super-sampling for images and videos~\cite{inrimagesupersampling, inrvideosupersampling}. More importantly, INRs have emerged as a promising approach for learning and representing 3D content, including learning 3D neural radiance field (NeRF) from 2D images for novel view synthesis, combining with image diffusion models for 3D model generation, as well as the representation, interpolation and completion of 3D shapes~\cite{nerf, dreamfusion, deepsdf}. Given the promising advantages of INRs, they are expected to become an important format for representing and storing 3D visual data. As more and more 3D visual data are generated in this format, we need a way to store, organize, and retrieve them as required.

A key aspect of organizing and retrieving INRs in a data store involves defining a similarity measure between INRs. This enables the retrieval of any stored INR from a database or data store using a query, such as an image or another similar INR. For instance, accurate INR retrieval can facilitate finding a similar but more detailed model or alternative recommended models from a collection of AI-generated 3D models or reconstructed models of real scenes. Retrieval of similar INRs from a store can also be potentially used in content creation pipelines~\cite{nerf_unreal_engine}, for applications such as scene completion~\cite{retrieval_scene_completion} and shape enhancement~\cite{retrieval_shape_completion}. Prior research has explored methods to determine similarity and retrieve 3D models represented by traditional formats like point clouds~\cite{ret_pc_pointnet, ret_pc_pointnetpp, ret_pc_pointnext}, meshes~\cite{ret_mesh_meshcnn, ret_mesh_field_convolutions, ret_mesh_MeshWalker}, and voxel grids~\cite{ret_occ_OCNN, ret_occ_adaptiveOCNN}. These approaches typically involve encoding shapes into embeddings using deep neural networks, where the similarity between shapes is indicated by the cosine similarity of their embeddings. However, there is little research on determining similarity and enabling the retrieval of INRs in a data store.  

In this work, our goal is to design a method that accurately and efficiently determines the similarity between INRs that represent shapes in a data store. This enables accurate retrieval of similar shapes (represented as INRs) from an INR data store using a query INR, as well as the clustering and organization of INRs representing similar shapes in the data store.

There are several challenges in enabling the accurate retrieval of shapes represented as INRs. Firstly, INRs can have many different architectures. For example, INRs can be multilayer perceptrons (MLPs) with different activation functions~\cite{deepsdf, siren} or more commonly, a combination of MLPs and different types of spatial feature grids, such as octrees~\cite{nglod}, triplanes~\cite{EG3D}, and hash grids~\cite{instantngp}. The only prior works that enable determining similarity between INRs only support an MLP-based architecture~\cite{inr2vec,nf2vec}, which is less commonly used today due to their limitation in representation capacity and training speed. This raises a challenge in how to flexibly support MLP-only INRs, feature grid-based INRs, and possibly any future architectures the query/stored INRs may have. Secondly, INRs can represent similar shapes using different implicit functions. For example, to represent 3D shapes, signed distance function (SDF), unsigned distance function (UDF), and occupancy field (Occ) are common implicit functions that INRs can encode~\cite{deepsdf, udf_neurips_2020, mescheder2019occupancy}. Different implicit functions are used because of their different advantages and heterogeneity of the source 3D data. First, these functions are capable of capturing different information of the same shape. For example, as we will demonstrate in \cref{fig:car_vis}, a car can be represented as a multi-layered car using UDF, capturing more intricate details \emph{inside} the car. The same car can also be represented with an SDF, such a representation can facilitate the creation of watertight surfaces that clearly delineate the car's interior from its exterior. This characteristic of SDFs is particularly advantageous for applications requiring clear boundary definitions, such as voxelization and marching cubes~\cite{voxelization,marchingcubes}. Second, the heterogeneity of source data often dictates the choice of implicit function. For example, UDF INRs are most commonly used to encode point clouds~\cite{udf_neurips_2020,udf_neurips_2022}. Different shape generation methods can also produce different representations; Pix2Vox generates voxel grids~\cite{gen_pix2vox}, while other efforts like SDFusion and AutoSDF opt for SDF to represent shapes~\cite{gen_sdfusion,gen_autosdf}. Therefore, it is essential to develop a general method that is effective across a broad spectrum of INR architectures and implicit functions, a solution that has yet to be achieved in current literature.

\begin{figure*}[ht]
\centering
    \includegraphics[width=\linewidth, trim=0 4 0 5, clip]{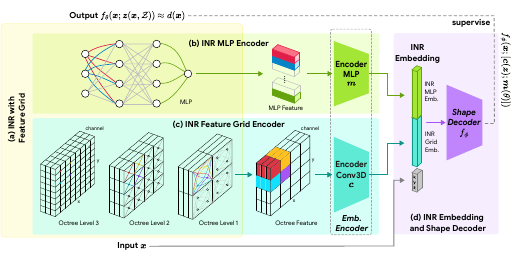}
    \caption{\small\textbf{Encoders trained to generate embeddings for grid-based INRs:} INRs with hash-table/octree/triplane based feature grids are used to generate embeddings for similarity calculations using encoders ($\bm{m}$ and $\bm{c}$).  The encoders take MLP weights and feature grid parameters as inputs to generate the INR Embedding. During training, the encoders ($\bm{m}$, $\bm{c}$) and the decoder ($f_\phi$) are jointly trained: the encoders to produce the INR Embedding and the decoder to reconstruct the original shape, ensuring the embeddings carry information about the shape. During inference, only the encoders are used to generate the INR Embedding for similarity calculations and retrieval.}
    \label{fig:inr_embed_creation}
\end{figure*}

In this work, we investigate different approaches to enable 3D shape retrieval and similarity estimation for INRs while addressing these challenges. We propose \name, a framework that enables identifying similar INRs and thus accurate retrieval of similar INRs given a query INR. \name flexibly supports INRs that use any architecture as well as any implicit function. We also compare \name with retrieval using existing approaches for traditional representations (for example, point clouds and multi-view images) by simply converting the shape INRs into these representations for retrieval. 

With \name, we determine the similarity between INRs by first generating embeddings for each INR. The similarity between these INRs is then estimated by calculating the cosine similarity of their embeddings. We now describe how we generate embeddings that are general across different INR architectures and implicit functions, allowing for comparison in a common embedding space.

First, we design an INR Embedding Encoder (hereafter referred to as \textit{Emb.\,Encoders}) that generates an embedding from the weights of the INR MLP and the learned parameters of the INR's feature grid. The key idea behind \textit{Emb.\,Encoder} is to encode the MLP and feature grid of the INR using an MLP encoder and a Conv3D encoder respectively. This encoder supports both MLP-only INRs and INRs with feature grids. The Conv3D encoder can take in learned feature vectors from any spatial grid as input. In this work, we demonstrate this generality with octree grids (NGLOD), triplanes (EG3D), and multi-resolution hash grids (iNGP)~\cite{nglod, EG3D, instantngp}. The embeddings created from these encoders are then concatenated to create the INR embedding for shape retrieval. To train these encoders, we also simultaneously train a \textit{Shape Decoder} that receives the INR embedding as input and outputs the INR's implicit function. The decoder is trained to approximate the implicit function values of the INRs used for training.

Second, to support retrieval of different implicit functions, we use separate \textit{Emb.\,Encoder} to encode INRs with different implicit functions. They are however trained to generate embeddings that are consistent across different implicit functions. To do this, during the encoder training process, we generate INRs and the corresponding INR embeddings, representing UDF, SDF, and Occ for \emph{each} training 3D shape. We then introduce two key regularization techniques to unify the embedding space. The first technique involves applying an explicit L2 Loss to reduce the discrepancies between embeddings of INRs that (despite their different implicit functions) represent the same shape. The second regularization is to use a \textit{Unified Shape Decoder} that outputs a single type of implicit function value (such as UDF) for all three implicit functions. The \textit{Shape Decoder} is only used to assist the training of \textit{Emb.\,Encoders}, by using a \textit{Unified Shape Decoder}, we improve the generality of the INR embedding created across different implicit functions. Our findings indicate that the \textit{Unified Shape Decoder} approach greatly contributes to unifying the latent space. We show that applying both regularizations is essential for ensuring the high retrieval accuracy of INRs across different implicit functions.

We demonstrate the effectiveness of our solution on the ShapeNet10 and Pix3D datasets. Compared to existing methods that perform retrieval of INRs directly, our method enables retrieval of INRs with feature grids, which cannot be done with existing solutions. Our method achieves 10.1\% higher retrieval accuracy on average than existing methods that can only retrieve shapes represented by MLP-only INRs~\cite{inr2vec}. We show that our regularization techniques enable retrieval of INRs across different implicit functions, achieving accuracy close to retrieval of INRs with the same implicit functions. Compared with retrieval methods applied to other representations such as point cloud and multi-view images converted from INRs, INRet achieves 12.1\% higher accuracy. Except in cases where INR architecture conversion is required, INRet has lower retrieval latency as it avoids the computation overhead associated with converting to other representations.

\noindent The contributions of this work are summarized as follows:

\hspace{0.1em}$\bullet$ We pose the challenge of evaluating similarity between INRs for the retrieval and organization of shape INRs in a data store. This involves assessing various techniques, including conversion to traditional formats and direct embedding creation from INRs.

\hspace{0.1em}$\bullet$ We propose a method to create embeddings from INRs, with or without feature grids, to represent shapes for retrieval and similarity evaluation purposes.

\hspace{0.1em}$\bullet$ We propose regularization techniques to produce embeddings in a unified latent space that facilitates comparison and retrieval across INRs with different implicit functions.

\hspace{0.1em}$\bullet$ We achieve higher retrieval accuracy on both the ShapeNet10 and Pix3D datasets compared to existing INR retrieval methods and methods involving conversion to traditional representations.

\section{Background \& Related Works}

\subsection{Implicit Neural Representation for Shapes}
Traditionally, 3D shapes have been represented with explicit representations including meshes, point clouds, and 3D voxels. Implicit Neural Representation (INR) has emerged as a novel paradigm for encapsulating shapes, employing neural networks to encode functions that implicitly represent a shape's surface. Seminal works like DeepSDF and Occupancy Networks demonstrate the feasibility of employing neural networks to encode signed distance function (SDF) or occupancy of 3D shapes \cite{deepsdf, mescheder2019occupancy}. Recent advancements extended this approach to encode unsigned distance function (UDF), showcasing higher representation quality for thinner surfaces \cite{sal, sald, udf_neurips_2020, udf_neurips_2022}.

\textbf{INRs with Multi-layer Perceptron.} Earlier works in INRs for shapes use simple multi-layer perceptrons (MLPs) with ReLU activations to represent the implicit functions \cite{deepsdf, mescheder2019occupancy, imnet, sal, sald, udf_neurips_2020, udf_neurips_2022}. SIREN proposed to use sinusoidal activation functions in MLPs to more efficiently encode higher frequency details~\cite{siren}. Since the implicit function is encoded in a single MLP, the MLP is usually relatively big and expensive to evaluate. The training of these MLPs to accurately represent the shapes is also time-consuming.

\textbf{INRs with Spatial Feature Grids.} While overfitting a large MLP to a shape can be difficult and computationally expensive, recent INRs for shapes use a combination of smaller MLPs and feature grids with learnable parameters. Peng et al. introduced Convolutional Occupancy Networks, which combine a trainable 3D dense feature grid and an MLP ~\cite{Peng2020ConvolutionalON}. Recent works have extended this notion and applied multi-level feature grids to encode and combine information at varying levels of detail. These multi-level spatial grids can be represented as sparse octrees as seen in NGLOD, VQAD, NeuralVDB, and ROAD \cite{nglod, vqad, NeuralVDB, road}. EG3D and iNGP introduced the idea of using triplanes multi-level hash grids to store these features at a fixed memory budget~\cite{EG3D, instantngp}. Sparse octrees, triplanes, and hash grids have seen wide applications in implicit neural representations for shapes or for radiance fields \cite{nerfstudio,tensorrf,xie2022neural}. Compared to INRs with only MLP, they significantly improve representation quality, as well as training and rendering speed. Our method considers INRs with or without the spatial grid for retrieval. We do so by optionally encoding the spatial grid for the INR embedding creation.

\subsection{Shape Retrieval}
\textbf{INR Retrieval.}
Numerous techniques have been developed to encode 3D shapes using INRs. The seminal work DeepSDF employs a shared Multi-Layer Perceptron (MLP) with varying latent codes to represent distinct shape instances~\cite{deepsdf}. These latent codes can be used for shape retrieval, as akin shapes tend to exhibit similar codes. Nonetheless, the adoption of the shared MLP concept in subsequent research has been limited due to its compromised representation quality when contrasted with employing a dedicated MLP for each shape~\cite{overfit}. The retrieval of dedicated MLP INRs has been studied in~\cite{inr2vec}. However, MLP-only INRs still fall behind INRs with feature grids in terms of representation quality and training speed. The method allowing the retrieval of INRs with spatial feature grids and/or across INRs with different implicit functions has yet to be developed.

\textbf{Retrieval by Converting to Traditional Representations.}
Shape retrieval for traditional 3D representations has many established works with techniques proposed for voxels, meshes, point clouds, and multi-view images~\cite{ret_occ_OCNN, ret_mesh_MeshWalker, ret_pc_pointnet, ret_proj_view_gcn}. However, these methods do not directly apply to the retrieval of INRs. A viable approach to retrieve INRs for shapes is to first transform these representations into one of the aforementioned traditional representations, and then apply established retrieval methods. For comparison with our method, We select two representations: point clouds and multi-view images, as they achieve higher accuracy in retrieval compared to other traditional representations ~\cite{ret_occ_OCNN, ret_mesh_MeshWalker}. Besides higher accuracy, point-based and multi-view image-based methods also avoid the computational overhead of the voxel-based methods and the requirement for watertight surfaces for the mesh methods ~\cite{ret_occ_adaptiveOCNN, ret_mesh_field_convolutions}.

We use the state-of-the-art methods PointNeXt and View-GCN as point-based and multi-view images-based baselines for comparison~\cite{ret_pc_pointnext,ret_proj_view_gcn}.

\section{Methods}
\label{sec:methods}

\subsection{Preliminary - INR for Shapes}
In this section, we introduce the different INR implicit functions and architectures we consider in this work. Consider a general distance or occupancy function $d(\cdot)$, defined for input coordinates $\bm{x} \in \mathbb{R}^3$ on the input domain of $\mathrm{\Omega} = \{\|\bm{x}\|_\infty \leq 1 \rvert \bm{x} \in \mathbb{R}^3\}$. The goal of INR for shape is to approximate $d(\cdot)$ by a function $f_\theta$ parameterized by a neural network.
\begin{equation} \label{eq:f}
  f_\theta(\bm{x})\approx d(\bm{x}), \,\forall \bm{x}\in \mathrm{\Omega}.
\end{equation}
Popular choices for the implicit function include signed distance function (SDF, $d_s(\cdot)\!\in\!\mathbb{R}$), unsigned distance function (UDF, $d_u(\cdot)\!\in\!\mathbb{R}^+$), and occupancy fields (Occ, $d_o(\cdot)\!\in\!{\{-1, 1\}}$)~\cite{deepsdf, udf_neurips_2020, mescheder2019occupancy}.
INRs are trained to minimize the difference between $f_\theta(\bm{x})$ and $d(\bm{x})$. 
Earlier works parameterize the function with a multi-layer perceptron (MLP). 
More recent works combine a feature grid with a smaller MLP, where the MLP takes the feature $\bm{z}$ sampled from the feature grid $\mathcal{Z}$ as input.
\begin{equation} \label{eq:decode_to_distance_value}
  f_\theta(\bm{x}; \bm{z}(\bm{x}, \mathcal{Z}))\approx d(\bm{x}), \,\forall \bm{x}\in \mathrm{\Omega}.
\end{equation}
The feature grid $\mathcal{Z}$ has various forms including sparse voxel octree, triplane, and hash grids, for which we all consider in this work~\cite{nglod,EG3D,instantngp}. All of these grids can be multi-level, encoding features at different spatial resolutions. For a multi-level feature grid, at each level $l \in \{1, ..., L\}$, the feature vector $\bm{\psi}(\bm{x}; l, \mathcal{Z})$ is interpolated (i.e., trilinearly) from local features. The final feature vector $\bm{z}$ from the grid is a summation (octree, triplane) or concatenation (hash grid) of features from all levels. The feature vector is then optionally concatenated with the input coordinate $\bm{x}$ and fed to a shallow MLP to calculate the distance or occupancy value.

\subsection{Embedding Creation for INR with Feature Grids}
\label{method:embed_feature_grid}
We determine the similarity between 3D shapes represented as INRs by converting each INR into an embedding, and the similarity between shapes is determined by the cosine similarity between these embeddings. 
We demonstrate our process for creating embeddings from INRs with feature grids in \cref{fig:inr_embed_creation}. 
Given a trained INR with an MLP component parametrized by $\theta$ and a feature grid $\mathcal{Z}$, we use an MLP Encoder $\bm{m}$ and Conv3D Encoder $\bm{c}$ to encode the features from the MLP and feature grid components of the INR respectively. Collectively, the MLP encoder $\bm{m}$ and Conv3D Encoder $\bm{c}$ constitutes the \textit{Emb.\,Encoder}. If the INR only contains an MLP component, we can simply omit the Conv3D Encoder $\bm{c}$. For the INR MLP component, the flattened weights of INR's MLP become input vectors to the MLP encoder. The structure of the encoder MLP is described in App.~\ref{app:inr_mlp_encoder_arch}.

For the INR feature grid, we sample $(2N)^3$ feature vectors at a fixed resolution from the feature grid, i.e. $S = \{[x_1  x_2  x_3]^T | x_i = \pm (\frac{1}{2N} + \frac{n}{N}), \forall n \in \{1, 2, \dots, N-1\}\}$. The sampled features are used as inputs to the Conv3D encoder, a 3D convolutional network that fuses discrete spatial features with gradually increasing perception fields (see Appendix \ref{app:inr_conv3d_encoder_arch} for more details). We use an octree (visualized in 2D) as an example in Figure \ref{fig:inr_embed_creation}\textbf{(c)}. Depending on the sampling resolution and resolution of the octree level, the feature is either collected directly from the corners of the voxels (Octree Level 3 in the example), or interpolated using features stored at the corners of the voxel containing the sampling location (Octree Level 2 \& 1). The features collected from each level are summed together, simply adding zero if a voxel is missing (due to sparsity in the octree). The collected features are fed to a Conv3D Encoder to create the INR Grid Embedding. A similar summation process can be done by traversing through the triplane grid levels. For a multi-resolution hash grid-based INR, we retrieve the features directly at the sampled locations using the original hash function and hash table. The MLP embedding and grid embedding are then concatenated to create our INR embedding.

\textbf{Training \textit{Emb.\,Encoders.}} During the encoder training process, we feed the concatenation of the INR embedding and the input coordinate $\bm{x}$ to the \textit{Shape Decoder} $f_\phi$. The decoder is supervised to generate the original implicit function that represents the shape.
Thus, the encoders are trained to generate embeddings that can be used to regenerate the original shape using the decoder. 
The following equation describes the process, where the decoder $f_\phi$ approximates the implicit function value of the original shape:
\begin{equation} \label{eq:decoder_approx}
  f_\phi(\bm{x}; [\bm{c}(\bm{z}); \bm{m}(\theta)])\approx d_{i\in{s,u,o}}(x)[\approx f_\theta(\bm{x}; \bm{z}(\bm{x}, \mathcal{Z}))].
\end{equation}
Note that since the INR parametrized by $\bm{\theta}, \bm{z}$ also encodes the implicit function, the INR Embedding $[\bm{c}(\bm{z}); \bm{m}(\theta)]$ is trained to contain information of the original shape.

\begin{figure*}[ht]
\centering
    \includegraphics[width=\linewidth]{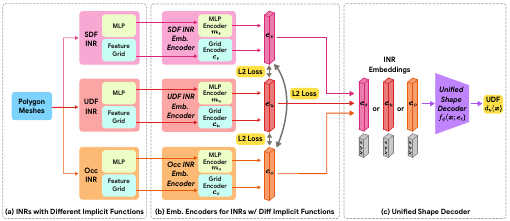}
    \caption{\textbf{INR Embed.\,Creation for INRs with Different Implicit Functions.} (a) For each shape, we train INRs with different implicit functions. (b) We train different encoders for INRs with different implicit functions. The differences between embeddings created by the encoders are minimized by L2 loss. (c) We feed the embeddings into a \textit{Unified Shape Decoder} to recreate the UDF of the original shape.}
    \label{fig:inr_cross_implicit}
\end{figure*}

\textbf{Supporting other INR architectures.}
\label{method:embed_diff_arch}
\name assumes a separate encoder for each type of INR architecture that is supported. Our proposed encoders already support the commonly used octree-based, triplane, and hash grid INR architectures. A similar feature grid sampling approach can be used to also train an encoder for any new grid-based architecture. Alternatively, other architectures can still be used with the above two encoders by using a distillation technique that converts a new INR architecture into one of the representations that we support. We describe how this can be done in App.~\ref{app:distillation}.

\subsection{Unified Latent Space for INRs with different Implicit Functions}
\label{method:embed_diff_implicit}
Besides different architectures, INRs can encode different implicit functions for the same underlying shape. To support multiple implicit functions of the same shape, we train separate encoders for each implicit function. To ensure that the generated embeddings map to the unified latent space, we apply two regularization techniques during the encoder training process.

The first regularization applied is explicit L2 loss to minimize the difference between embeddings created from INRs for different implicit functions of the same shape. The second regularization is to use a \textit{Unified Shape Decoder} that outputs a single type of implicit function value (such as UDF) for all three implicit functions. We show in App.~\ref{app:shape_decoder_implicit_choice} that the specific choice of \textit{Unified Shape Decoder} implicit function (UDF \textit{vs.} SDF \textit{vs.} Occ) has minimal impact on the retrieval accuracy. The key is that both regularizations are applied.

The overall loss function for this process is
\begin{equation}
\mathcal{L} = \sum_{i \in \{s, u, o\}}|f_\phi(\bm{x}; \bm{e}_i)-d_u(\bm{x})| + \lambda \sum_{i, j \in \{s, u, o\}}\|\bm{e}_i - \bm{e}_j\|^2
\end{equation}

$\bm{e}_i = [\bm{c}_i(\bm{z}_i); \bm{m}_i(\theta_i)]$ is the INR embedding for the implicit function $i$ (unsigned/signed distance or occupancy). The first part of the loss is the difference between the \textit{Unified Shape Decoder's} output with the groundtruth unsigned distance $d_u$, and the second part is the L2 loss between the INR embeddings. $\lambda$ is a hyperparameter balancing the contribution of the two parts, we found a $\lambda$ of 1 works well in practice. During the encoder training process, we create INRs for all implicit functions of each training shape to train the encoders to generate embeddings that share the unified latent space. 

\section{Retrieval by Converting to Explicit Representations}
\label{method:embed_other_rep}
An alternative approach to evaluate similarity and enable retrieval of similar INRs is to first convert to an explicit representation, such as point clouds or multi-view images. This approach would enable the use of prior research to evaluate similarity between shapes represented in these traditional formats. In this work, we also evaluate the effectiveness of this approach in comparison to directly using INR embeddings. 
Conversion to point clouds and multi-view images from SDF INRs can be done through spherical tracing~\cite{spheretrace}. 
For point cloud sampling, we start spherical tracing from randomly selected locations and directions until enough points on the surface of the object are collected~\cite{nglod}. 
The multi-view images are also collected via spherical tracing starting from camera centers at fixed positions. 
For UDF INRs, we use the damped spherical tracing presented in prior work~\cite{udf_neurips_2020} that avoids overshooting. 
For the occupancy values, spherical tracing is not possible so we follow the method presented in Occupancy Networks~\cite{mescheder2019occupancy}. Using occupancy values sampled at fixed resolutions from the trained INR, 
we combine isosurface extraction and the marching cubes algorithm to create the surface mesh of the object~\cite{marchingcubes}. We then perform point cloud sampling and multi-view image rendering from the constructed mesh. To generate embeddings for similarity evaluations from these formats, we use PointNeXt~\cite{ret_pc_pointnext} for extracted point clouds, and View-GCN~\cite{ret_proj_view_gcn} for multi-view images (details are in App.~\ref{app:pointnext_viewgcn}).


\section{Evaluation}
\label{sec:evaluation}

\subsection{Experimental Setup}
\label{sec:setup}

\begin{table*}[]
\centering
\begin{tabular}{c|c|c|c|c|c|c}
\hline
Method & \multicolumn{3}{c|}{Ours} & inr2vec & PointNeXt & View-GCN  \\ \hline
Input Type       & {NGLOD}      & {EG3D}      & iNGP     & MLP INR         & Point Cloud      & Multi-View \\ \hline
mAP @ 1          & 82.6/\underline{74.3}  & \underline{82.8}/74.1          & \textbf{84.2}/\textbf{78.0}            & 73.4/71.4          & 71.2/69.3           & 73.6/70.5               \\ 
Ret. Speed(s)  & {0.034} & {0.031}         & 0.14             & 0.062          & 0.98            & 3.05               \\ \hline
\end{tabular}
\caption{Shape Retrieval Accuracy and Speed on SDF INRs (ShapeNet10/Pix3D)}
\label{tab:sdf_retrieval}

\centering
\begin{tabular}{c|c|c|c|c|c}
\hline
Method & \multicolumn{3}{c|}{Ours} & PointNeXt & View-GCN  \\ \hline
Input Type       & NGLOD      & EG3D      & iNGP         & Point Cloud      & Multi-View Images \\ \hline
mAP @ 1          & 76.2/\underline{71.3}          & \underline{76.4}/70.4          & \textbf{79.2}/\textbf{75.5}            & 70.2/67.1          & 71.4/68.2               \\ 
Ret. Speed(s)  & 30.2          & 34.1          & 29.6             & 1.26          & 4.57                \\ \hline
\end{tabular}
\caption{Shape Retrieval Accuracy with MLP-only INR as Query (ShapeNet10/Pix3D)}
\label{tab:cross_architecture}
\end{table*}

\noindent{\bf Datasets.}
We use ShapeNet and Pix3D to demonstrate the generality and robustness of our solution~\cite{shapenet,pix3d}. 
For the ShapeNet10 dataset, each category has 50 models for training and 50 models for testing. 
For Pix3D, we use 70\% of the shapes from each category as training data and 30\% as testing data.

\noindent{\bf Metrics.} We evaluate the effectiveness of our framework in identifying similar shapes in the data store by using a test INR shape to retrieve the most similar $k$ INR shapes. We report the mean Average Precision (mAP) as the average accuracy of retrieving a shape from the same category as the query shape across all shapes in the test set. We also report precision, recall, and F1 score as defined in the ShapeNet retrieval challenge in the App.~\ref{app_sec:exp}~\cite{shrec16}. 

\noindent{\bf Baselines.}
We compare against inr2vec for retrieval from INRs by directly encoding the INR weights. We also compare with PointNeXt and View-GCN by converting the trained INR to point-cloud and multi-view images, respectively.

\noindent{\bf Ablations.}
We only report key results in this section, and provide additional results and more detailed ablation studies in App.~\ref{app_sec:exp}.

\subsection{INR Retrieval with Feature Grids}
\label{exp:exp_grid}

To create a baseline shape INR data store, we train NGLOD (octree), EG3D (triplane) and iNGP (hash-grid) INRs with SDF to encode shapes from ShapeNet10 and Pix3D datasets~\cite{nglod,EG3D,instantngp}. The encoders are trained on our training set and used to generate embeddings for the test set. \cref{tab:sdf_retrieval} presents mAP@1 and retrieval speed (in seconds), and additional metrics are available in Appendix \ref{app_subsec:exp1}. Our comparison includes INRet against inr2vec, which performs retrieval on MLP-only INRs of the same implicit function. Additionally, we compare with PointNeXt and View-GCN by converting the trained iNGP INR to point clouds and multi-view images for retrieval.

As seen in~\cref{tab:sdf_retrieval}, INRet achieves the highest accuracy: on average 12.0\%, 15.4\%, and 12.6\% higher accuracy than inr2vec, PointNeXt and View-GCN methods respectively (for iNGP INRs). For \name, retrieving from iNGP INRs achieved slightly higher performance than retrieving from NGLOD and EG3D INRs. In terms of retrieval speed, \name on NGLOD and EG3D are the fastest, followed by inr2vec, which is slightly slower due to the large number of weights in the INR MLP. Compared to NGLOD and EG3D, from which the embedding can be directly summed from the feature grid, embedding sampling from the iNGP hash-grid is slower due to the higher overhead of the hash operations during sampling. 
Converting to point clouds or multi-view images for retrieval with PointNeXt or View-GCN is 1-2 orders of magnitude slower than directly encoding the INR weights.

In summary, \name enables high-accuracy retrieval of similar shape INRs. Converting to images or point clouds leads to lower accuracy due to information loss during the conversion process and incurs the latency overhead for format conversion. 

\subsection{INR Retrieval with Different Architectures}
\label{result:diff_arch}
In this section, we evaluate INRet's effectiveness in retrieving shapes across different INR architectures. Given an MLP-only INR as the query, we want to retrieve from a data store of INRs with different architectures from the query INR. We consider an MLP INR similar to that used by inr2vec as input. We apply the INR distillation technique discussed in \cref{method:embed_diff_arch} (Supporting other INR architectures) to convert the MLPs into INRs with feature grids to retrieve NGLOD, EG3D or iNGP INRs. 

As depicted in \cref{tab:cross_architecture}, following INR distillation, \name achieves an average accuracy of 73.8\%, 73.4\% and 77.4\% respectively for NGLOD, EG3D and iNGP \textit{Emb. Encoders} across the two datasets, surpassing the average accuracy of 72.4\% achieved by inr2vec. Our method also performs better than converting to point cloud or multi-view images. This highlights the robustness of our approach in adapting to different architectures not directly supported by the encoder. Despite performing a distillation, the converted INRs with a feature grid can be used to generate better embeddings for retrieval when compared with generating the embeddings directly from MLP-only INR. While format conversion introduces some overhead (approximately 30 seconds), a potential speed-accuracy tradeoff could be explored by converting to point clouds/images when \name lacks a pre-trained encoder for a new architecture.

\subsection{INR Retrieval with Different Implicit Functions}
\label{ssec:inr_diff_imp}
In this section, we evaluate the effectiveness of our method in performing INR retrieval across INRs with different implicit functions (i.e., UDF, SDF and Occ). We compare against inr2vec and point-based and image-based methods.

\begin{table}[]
\centering
\setlength{\tabcolsep}{6pt} 
\begin{tabular}{c@{\hspace{1pt}}|@{\hspace{1pt}}c@{\hspace{1pt}}|@{\hspace{1pt}}c@{\hspace{1pt}}|@{\hspace{1pt}}c@{\hspace{1pt}}|@{\hspace{1pt}}c}
\hline
\multicolumn{2}{c|}{} & \multicolumn{3}{c}{Retrieval INR} \\ \cline{3-5}
\multicolumn{2}{c|}{} & UDF & SDF & Occ \\ \hline
\multirow{4}{*}{\rotatebox[origin=c]{90}{Query}} & UDF & 
{\begin{tabular}[c]{@{}l@{}}\lA{80.2}/\lB{80.8}/\lC{83.0}\\ \lD{68.8}/\lE{72.0}/\lF{70.8}\end{tabular}} & 
{\begin{tabular}[c]{@{}l@{}}\lA{81.4}/\lB{79.4}/\lC{79.0}\\ \lD{10.4}/\lE{61.8}/\lF{72.6}\end{tabular}} & 
{\begin{tabular}[c]{@{}l@{}}\lA{78.8}/\lB{79.2}/\lC{80.4}\\ \lD{{\textcolor{white}0}8.8}/\lE{58.2}/\lF{68.4}\end{tabular}}  \\ \cline{2-5} 
& SDF & 
{\begin{tabular}[c]{@{}l@{}}\lA{82.2}/\lB{81.2}/\lC{81.8}\\ \lD{11.4}/\lE{62.2}/\lF{70.2}\end{tabular}} & 
{\begin{tabular}[c]{@{}l@{}}\lA{83.4}/\lB{82.4}/\lC{84.6}\\ \lD{70.2}/\lE{67.2}/\lF{69.4}\end{tabular}} & 
{\begin{tabular}[c]{@{}l@{}}\lA{79.2}/\lB{79.6}/\lC{82.4}\\ \lD{10.4}/\lE{56.2}/\lF{68.8}\end{tabular}} \\ \cline{2-5} 
& Occ & 
{\begin{tabular}[c]{@{}l@{}}\lA{76.0}/\lB{79.8}/\lC{81.0}\\ \lD{{\textcolor{white}0}9.2}/\lE{55.4}/\lF{62.6}\end{tabular}} & 
{\begin{tabular}[c]{@{}l@{}}\lA{77.0}/\lB{79.4}/\lC{82.6}\\ \lD{10.4}/\lE{56.2}/\lF{61.8}\end{tabular}} & 
{\begin{tabular}[c]{@{}l@{}}\lA{76.8}/\lB{80.0}/\lC{83.0}\\ \lD{69.4}/\lE{51.2}/\lF{66.4}\end{tabular}} \\
\hline
\multicolumn{2}{c|}{Average} & \multicolumn{3}{c}
{\begin{tabular}[c]{@{}l@{}}\lA{79.4}/\lB{\underline{80.2}}/\lC{\textbf{82.0}}\\ \lD{29.9}/\lE{60.0}/\lF{67.9}\end{tabular}} \\
\hline
\multicolumn{2}{c|}{\textit{Legend}} & \multicolumn{3}{c}
{\begin{tabular}[c]{@{}l@{}}\textit{Ours}: \hspace{14pt}\lA{NGLOD}/\hspace{8.5pt}\lB{\underline{EG3D}}\hspace{8.5pt}/\hspace{10pt}\lC{\textbf{iNGP}}\\ \textit{Baselines}: \lD{inr2vec}/\lE{PointNeXt}/\lF{View-GCN}\end{tabular}} \\
\hline
\end{tabular}
\caption{Shape Retrieval Accuracy on Different Implicit Functions INRs for ShapeNet10}
\label{tab:cross_retrieval}
\end{table}

\begin{table}[]
\centering
\setlength{\tabcolsep}{3pt} 
\begin{tabular}{c@{\hspace{2pt}}|@{\hspace{2pt}}c@{\hspace{2pt}}|@{\hspace{2pt}}c@{\hspace{2pt}}|@{\hspace{2pt}}c@{\hspace{2pt}}|@{\hspace{2pt}}c}
\hline
\multicolumn{2}{c@{\hspace{2pt}}|}{} & \multicolumn{3}{c}{Retrieval INR} \\ \cline{3-5}
\multicolumn{2}{c@{\hspace{2pt}}|}{} & UDF & SDF & Occ \\ \hline
\multirow{3}{*}{\rotatebox[origin=c]{90}{Query}} & UDF & 83.4/83.4/83.0 & {\textcolor{white}0}9.4/52.4/79.0 & 10.8/51.8/80.4 \\
& SDF & 10.8/57.8/81.8 & 82.4/81.4/84.6 & {\textcolor{white}0}9.6/53.2/82.4 \\
& Occ & 11.4/65.4/81.0 & 10.2/53.2/82.6 & 81.6/82.4/83.0 \\ \hline
\multicolumn{2}{c@{\hspace{2pt}}|}{Average} & \multicolumn{3}{c}{34.0/\underline{64.6}/\textbf{82.0}} \\
\hline
\end{tabular}
\caption{Shape Retrieval Accuracy on iNGP INRs for ShapeNet10 ({No Reg.} / \underline{L2 Reg.} / \textbf{L2 Reg.} \& \textbf{\textit{Unified Shape Decoder}})}
\label{tab:cross_retrieval_regularization_ablation}
\end{table}

As seen in \cref{tab:cross_retrieval}, using our method to retrieve iNGP INRs with different implicit functions achieves the highest 82.0\% accuracy, which is higher than the accuracy achieved with inr2vec, PointNeXt, and View-GCN. In particular, inr2vec achieves very low accuracy (around 10\%) for retrieving INRs with different implicit functions. As seen in \cref{tab:cross_retrieval_regularization_ablation}, using INRet to retrieve iNGP with different implicit functions also achieves very low accuracy (around 10\%) if no regularization is used. The average accuracy for retrieval improves significantly if the L2 regularization (64.6\% accuracy) and both regularizations (82.0\% accuracy) are applied. 

In this section, we used the original meshes to sample SDF, UDF and Occ values to train the INRs with different implicit functions. However, for the UDF INRs, one can also train the INRs given an input point cloud, we demonstrate in Appendix \ref{app_sec:udf_pc} that the retrieval accuracy does not change significantly compared to when the UDF INRs are trained using the meshes. This demonstrates that by enabling retrieval from INRs with different implicit functions with \name, we ultimately enable retrieval of INRs trained with different 3D input modalities.

\subsection{Retrieval Visualization}
\begin{figure}[h]
\centering
    \includegraphics[width=\linewidth, trim=30 0 30 0, clip]{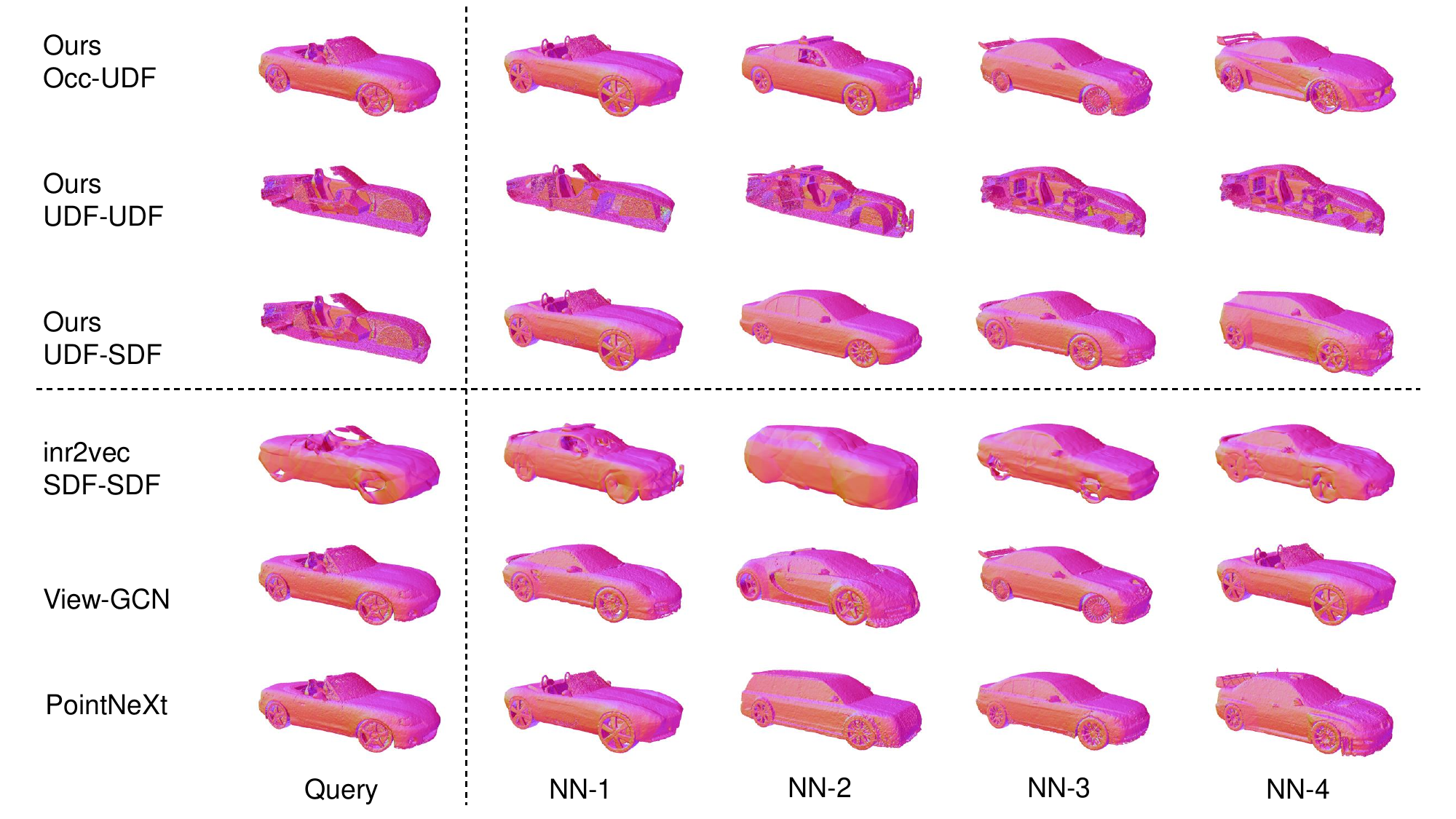}
    \caption{Retrieval Qualitative Comparison.}
    \label{fig:car_vis}
\end{figure}

We visualize the retrieved shapes in \cref{fig:car_vis}. In particular, we demonstrate our solution enables the retrieval of INRs with different implicit functions, which is not possible with other baseline solutions.

\cref{fig:car_vis} shows the query and retrieved results for the car class in ShapeNet. As we can see from the figure, given a convertible as the query, our method consistently retrieves the other convertibles from the dataset as the top candidates while most other methods fail to do so. \cref{fig:car_vis} also demonstrates INRet's ability to retrieve watertight surfaces (represented with SDF INR) from surfaces with multiple inner layers (represented with UDF INR). For the "Ours UDF-UDF" and "Ours UDF-SDF" rows, we show renderings of the same query car, but with the middle cut open when the shape is represented using a UDF INR. We can see that the UDF INR can capture the details inside the car. When used as the query, these UDF INRs can retrieve cars represented with different implicit functions correctly. In addition, compared with the renderings demonstrated in the row "inr2vec SDF-SDF" which used an MLP-only INR to represent the underlying shape. Our method uses iNGP to represent the shape thus capturing more details. We provide additional quantitative results on the reconstruction quality in App.~\ref{app_ssec:recon_quality}.

While different implicit functions have very different values, we show in App.~\ref{app:udf_sdf_relation} that for the same underlying shape, their values are highly correlated.

\subsection{Embedding Space t-SNE Visualization}

\begin{figure}[h]
  \centering

  \begin{subfigure}[b]{0.23\textwidth}
    \includegraphics[width=\textwidth, trim=70 440 650 70, clip]{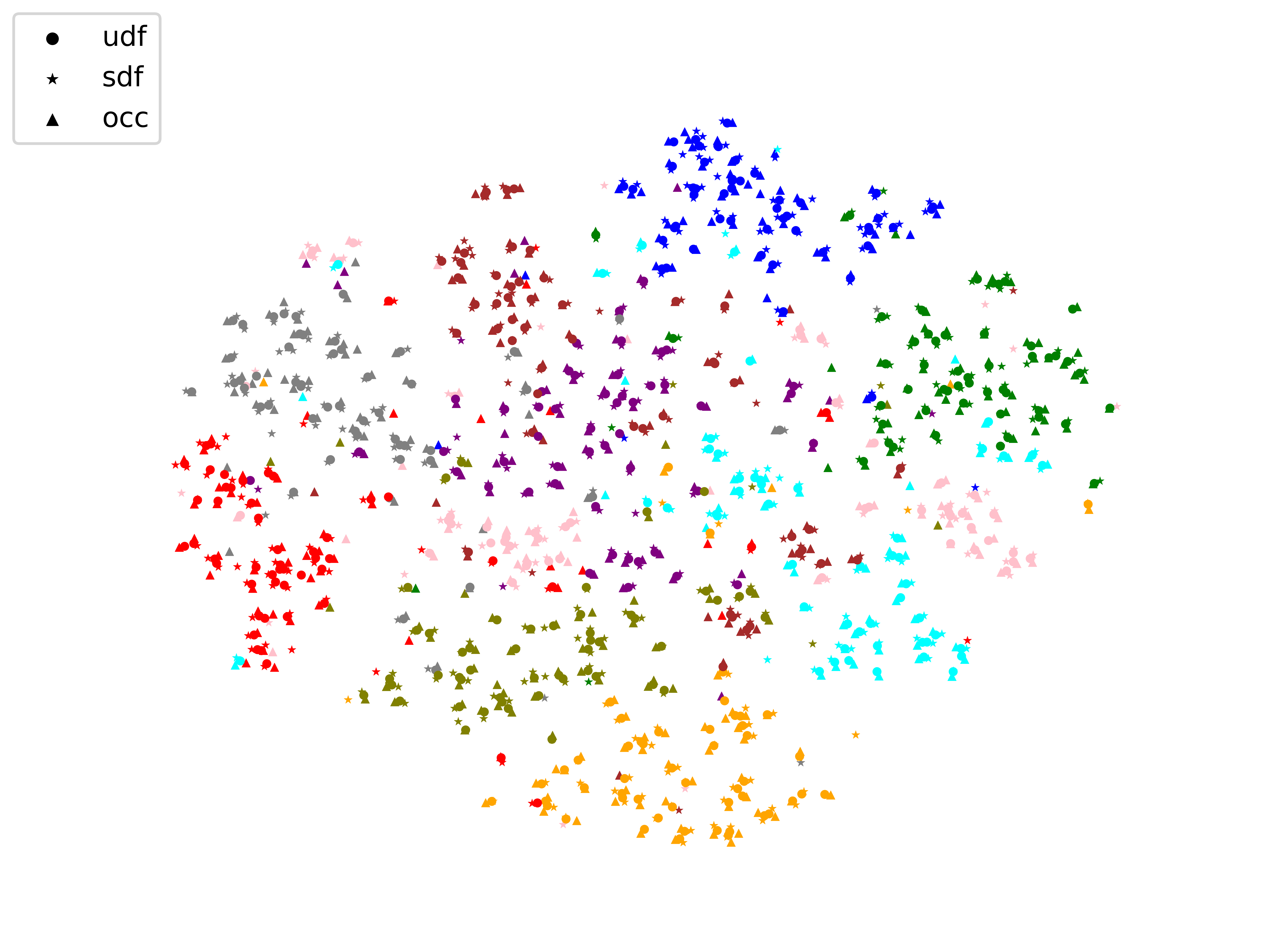}
    \caption{INR Emb.\,\,tSNE with different \textit{Shape Decoders}}
    \label{fig:tsne_mix_no_decoder}
  \end{subfigure}
  \hfill
  \begin{subfigure}[b]{0.23\textwidth}
    \includegraphics[width=\textwidth, trim=70 440 650 70, clip]{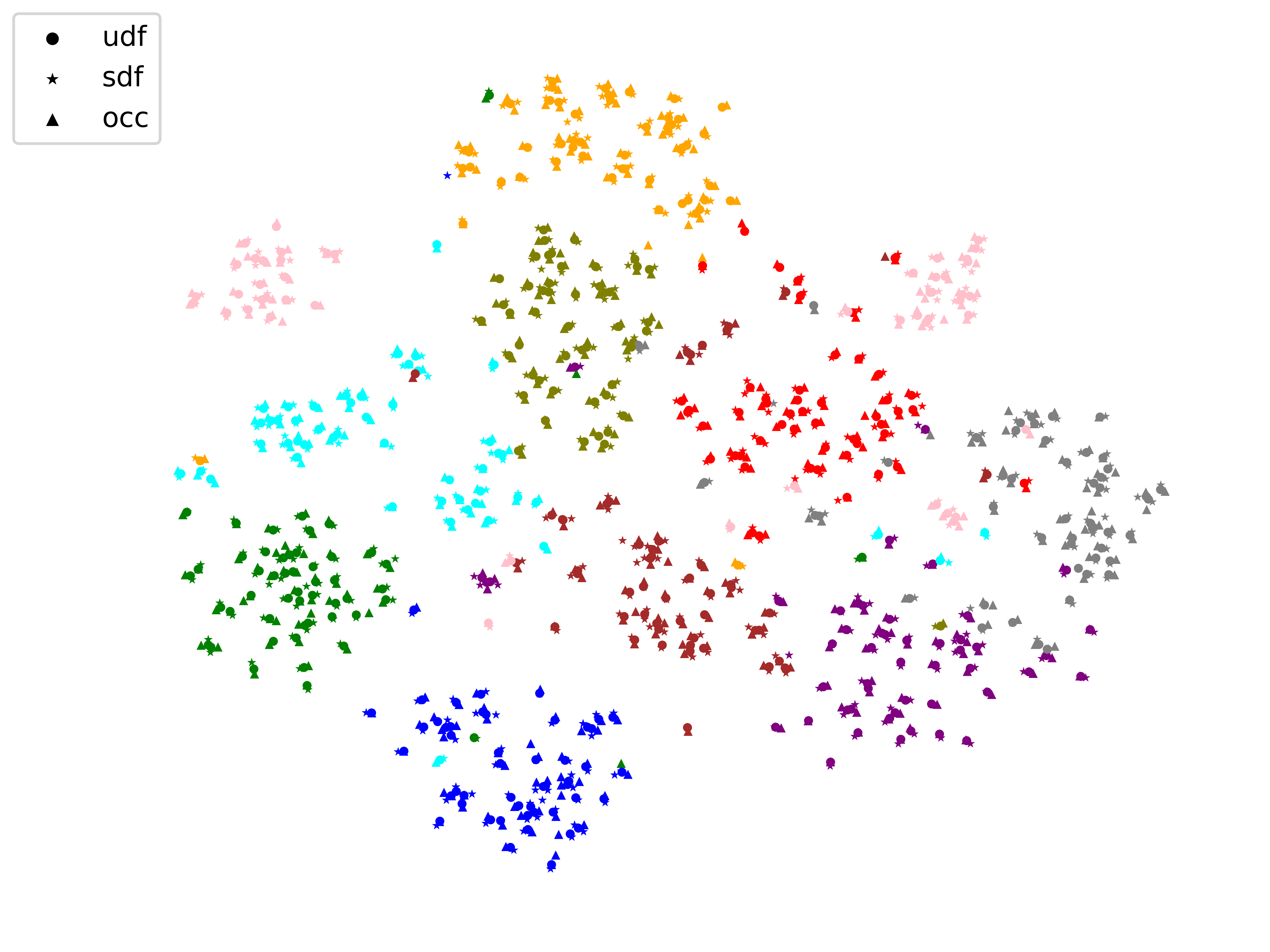}
    \caption{INR Emb.\,\,tSNE with \textit{Unified Shape Decoder}}
    \label{fig:tsne_mix_with_decoder}
  \end{subfigure}

  \caption{INR Embedding tSNE Plot}
  \label{fig:tsne_without_and_with_decoder}
\end{figure}

In \cref{fig:tsne_without_and_with_decoder}, we provide the t-SNE plot of the INR embeddings created by the \textit{Emb. Encoders} trained with and without the \textit{Unified Shape Decoder}. When trained using different \textit{Shape Decoders}, the embeddings for shapes belonging to the same category are much more spread out, this is likely due to the different decoders requiring the INR embedding for the same shape to be used for different purposes (decoding UDF, SDF, or Occ). This misses the regularization from the unified decoder that further minimizes the difference between embeddings of the same shape represented with INRs with different implicit functions.

\vspace{-10pt}
\section{Conclusion}
\label{sec:conclusion}
In this work, we presented a new framework for determining similarity between INRs that can be used for accurate retrieval of INRs from a data store. We proposed a new encoding method for INRs with feature grids including the octree and hash table based grids. By using L2 loss and a common decoder as regularizations, INRet also enables the retrieval of INRs across different implicit functions. On ShapeNet10 and Pix3D datasets, INRet demonstrates more than 10\% improvement in retrieval accuracy compared to prior work on INR retrieval and retrieval by conversion to point cloud and multi-view images. Compared to point cloud and multi-view image retrieval methods, INRet is also faster by avoiding the conversion overhead when retrieving INRs with same or different implicit functions.

\setlength{\textfloatsep}{1pt}
\setlength{\floatsep}{1pt}
\setlength{\dbltextfloatsep}{5pt}
\setlength{\dblfloatsep}{5pt}
\setlength{\abovecaptionskip}{3pt}
\setlength{\belowcaptionskip}{10pt}

{
    \small
    \bibliographystyle{ieeenat_fullname}
    \bibliography{main}
}
\clearpage
\setcounter{page}{1}
\maketitlesupplementary

\newpage
\section{Architecture and Training Details}
\label{app:arch}

\begin{figure*}
\centering
    \includegraphics[width=\linewidth]{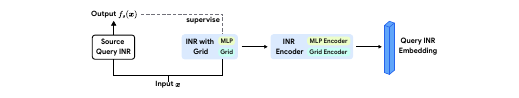}
    \caption{\textbf{INR Embedding Creation for INRs with Different Architectures} }
    \label{fig:inr_cross_arch}
\end{figure*}

\begin{figure*}
\centering
    \includegraphics[width=\linewidth]{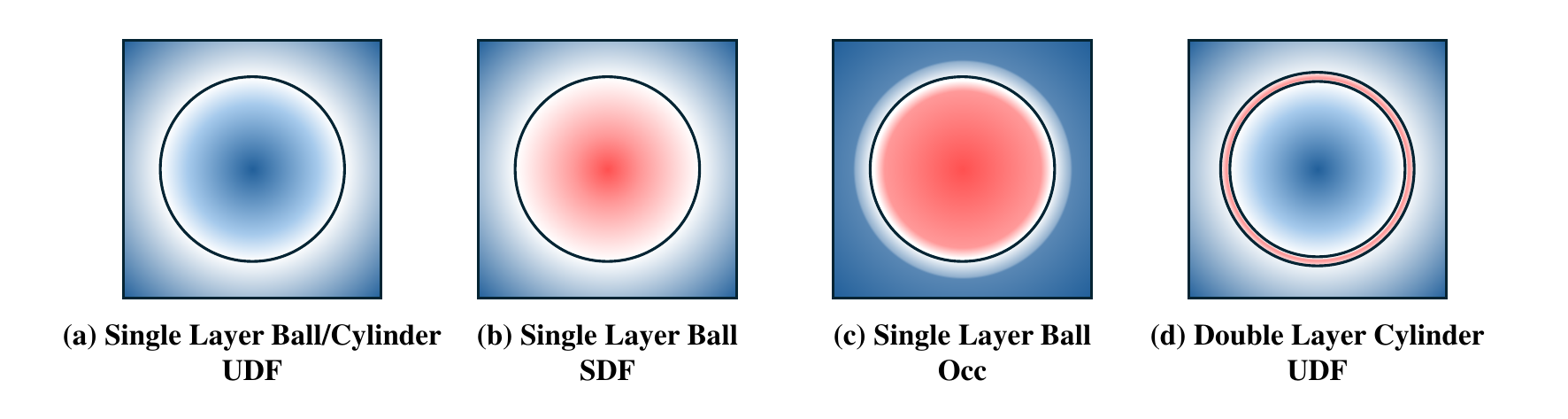}
    \caption{\textbf{Implicit Function Visual Representation for Cross Section of Different Shapes} }
    \label{fig:udfsdf}
\end{figure*}

\begin{table*}[h]
\centering
\begin{tabular}{c|cc|c|c|c}
\hline
Method     & \multicolumn{2}{c|}{Ours}  & inr2vec & PointNeXt & View-GCN   \\ \hline
Input Type & \multicolumn{1}{c|}{NGLOD} & iNGP  & MLP INR         & Point Cloud      & Multi-View Image \\ \hline
mAP @ 1    & \multicolumn{1}{c|}{\underline{82.6}}  & \textbf{84.2}  & 73.4            & 68.0             & 71.6             \\
mAP @ 5    & \multicolumn{1}{c|}{\textbf{94.4}}  & \textbf{94.4}  & 89.8            & 87.2             & 88.2             \\
mAP @ 10   & \multicolumn{1}{c|}{\underline{96.4}}  & \textbf{96.6}  & 92.4            & 89.6             & 90.4             \\
F1 @ 10     & \multicolumn{1}{c|}{\underline{80.8}} & \textbf{81.8} & 72.0           & 67.8            & 70.4            \\ \hline
\end{tabular}
\caption{Shape Retrieval Accuracy Metrics on ShapeNet10}
\label{tab:appendix_exp1_shapenet}

\centering
\begin{tabular}{c|cc|c|c|c}
\hline
Method     & \multicolumn{2}{c|}{Ours} & inr2vec & PointNeXt & View-GCN   \\ \hline
Input Type & \multicolumn{1}{c|}{NGLOD} & iNGP & MLP INR         & Point Cloud      & Multi-View Image \\ \hline
mAP @ 1    & \multicolumn{1}{c|}{\underline{76.5}}  & \textbf{78.0} & 71.4            & 66.3             & 68.5             \\
mAP @ 5    & \multicolumn{1}{c|}{\underline{88.9}}  & \textbf{93.8} & 91.0            & 81.5             & 87.2             \\
mAP @ 10   & \multicolumn{1}{c|}{92.6}  & \underline{94.3} & \textbf{95.5}            & 88.4             & 93.3             \\
P @ 10      & \multicolumn{1}{c|}{\underline{66.7}}  & \textbf{68.0} & 62.3            & 59.9             & 61.0             \\
R @ 10      & \multicolumn{1}{c|}{\underline{75.2}}  & \textbf{76.1} & 69.0            & 68.5             & 70.3             \\
F1 @ 10     & \multicolumn{1}{c|}{\underline{70.7}}  & \textbf{71.9} & 65.3            & 63.9             & 65.2             \\ \hline
\end{tabular}
\caption{Shape Retrieval Accuracy Metrics on Pix3d}
\label{tab:appendix_exp1_pix3d}

\centering
\begin{tabular}{c|c|ccc}
\hline
\multicolumn{2}{c|}{\multirow{2}{*}{}}                & \multicolumn{3}{c}{Retrieval INR}                                                                                   \\ \cline{3-5} 
\multicolumn{2}{c|}{}                           & \multicolumn{1}{c|}{UDF}                    & \multicolumn{1}{c|}{SDF}                     & Occ                     \\ \hline
\multirow{3}{*}{\rotatebox[origin=c]{90}{Query}} & UDF & \multicolumn{1}{c|}{\lA{69.4}/\lB{70.4}/\lC{66.7}/\lD{61.2}/\lE{68.5}} & \multicolumn{1}{c|}{\lA{71.6}/\lB{72.8}/\lC{12.2}/\lD{59.9}/\lE{66.4}} & \lA{71.6}/\lB{71.6}/\lC{10.7}/\lD{60.8}/\lE{61.2} \\ \cline{2-5} 
                           & SDF & \multicolumn{1}{c|}{\lA{67.9}/\lB{72.8}/\lC{12.4}/\lD{61.4}/\lE{67.4}} & \multicolumn{1}{c|}{\lA{74.1}/\lB{79.0}/\lC{71.5}/\lD{62.3}/\lE{67.2}} & \lA{67.9}/\lB{69.1}/\lC{11.9}/\lD{54.4}/\lE{59.7} \\ \cline{2-5} 
                           & Occ & \multicolumn{1}{c|}{\lA{69.1}/\lB{67.9}/\lC{13.1}/\lD{57.6}/\lE{60.4}} & \multicolumn{1}{c|}{\lA{72.3}/\lB{74.1}/\lC{11.8}/\lD{56.8}/\lE{60.9}} & \lA{68.9}/\lB{69.1}/\lC{65.4}/\lD{58.2}/\lE{61.3} \\ \hline
\multicolumn{2}{c|}{Average} & \multicolumn{3}{c}{\lA{\underline{70.3}}/\lB{\textbf{71.9}}/\lC{30.6}/\lD{59.2}/\lE{63.7}} \\
\hline
\multicolumn{2}{c|}{\textit{Legend}} & \multicolumn{3}{c}{\lA{\underline{NGLOD}} / \lB{\textbf{iNGP}} / \lC{inr2vec} / \lD{PointNeXt} / \lE{View-GCN}} \\
\hline
\end{tabular}
\caption{Shape Retrieval Accuracy for Different Implicit Function INRs on Pix3D}
\label{tab:cross_retrieval_pix3d}

\centering
\begin{tabular}{c|c|ccc}
\hline
\multicolumn{2}{c|}{\multirow{2}{*}{}}                & \multicolumn{3}{c}{Retrieval INR}                                                                                   \\ \cline{3-5} 
\multicolumn{2}{c|}{}                           & \multicolumn{1}{c|}{UDF}                    & \multicolumn{1}{c|}{SDF}                     & Occ                     \\ \hline
\multirow{3}{*}{\rotatebox[origin=c]{90}{Query}} & UDF & \multicolumn{1}{c|}{80.2/83.0}              & \multicolumn{1}{c|}{91.2($+${\textcolor{white}0}9.8)/88.4($+${\textcolor{white}0}9.4)}   & 86.6($+${\textcolor{white}0}7.8)/87.4($+${\textcolor{white}0}7.0)   \\ \cline{3-4}
                  & SDF & \multicolumn{1}{c|}{92.0($+${\textcolor{white}0}9.8)/93.2($+$11.4)} & \multicolumn{1}{c|}{83.4/84.6}               & 90.2($+$11.0)/92.8($+$10.4) \\ \cline{4-5}
                        & Occ & \multicolumn{1}{c|}{85.6($+${\textcolor{white}0}9.6)/89.4($+${\textcolor{white}0}8.4)}  & \multicolumn{1}{c|}{87.8($+$10.8)/92.6($+$10.0)} & 76.8/83.0               \\ \hline
\multicolumn{2}{c|}{\textit{Legend}} & \multicolumn{3}{c}{NGLOD ($+$Improvement) / iNGP ($+$Improvement)} \\
\hline
\end{tabular}
\caption{Shape Retrieval Accuracy for Different Implicit Function INRs on ShapeNet10, allowing Retrieval of Same Shape. In (bracket), we report the improvement in retrieval accuracy when retrieving the same shape is allowed.}
\label{tab:cross_implicit_retrieve_same_shape}
\end{table*}

\subsection{INR Architecture and Training Detail}
\label{app_ssec:inr_training_detail}

\noindent{\bf INR Training Losses.} We apply different loss functions for different implicit functions. For the signed distance function, we follow the method in~\cite{nglod}.
\begin{equation}
\mathcal{L}_s(f_\theta(\bm{x}), d_s(\bm{x})) = \|f_\theta(\bm{x}) - d_s(\bm{x})\|^2
\end{equation}
For the unsigned distance function, we follow the method in~\cite{udf_neurips_2020}
\begin{equation}
\mathcal{L}_u(f_\theta(\bm{x}), d_u(\bm{x})) = |f_\theta(\bm{x}) - d_u(\bm{x})|
\end{equation}
For the occupancy field, we also apply an L1 loss similar to the unsigned distance function.

\noindent{\bf INR Architectures.} We implement the INR architectures in this work using the NVIDIA Kaolin Wisp library ~\cite{KaolinWispLibrary}. We follow the default configurations for the NGLOD, iNGP and EG3D architectures. For NGLOD, we use an octree with 6 levels, but do not store features in the first 2 levels following the default configuration. We use a feature size of 8 for each level. For iNGP, we utilize 4 levels for the hash grid. The minimum and maximum grid resolutions are set to 16 and 512, respectively, with a maximum hashtable size of $2^{19}$ for each level. The feature size is 2 for each level. For EG3D, we use the default configuration with 1 level with a feature size of 8, but our solution could support multiple levels. For all grids, we initialize the grid features by sampling from a normal distribution with a mean of 0 and a standard deviation of 0.01. For the MLP-only INR, we follow the configuration in~\cite{inr2vec}, and train a SIREN INR with 4 hidden layers and 512 hidden nodes~\cite{siren}. The MLP uses sine activation functions.

\noindent{\bf INR Training.} We use the polygon meshes from the ShapeNet10 and Pix3D datasets to generate SDF, UDF and Occ values to train the INRs. We use the sampling method from Kaolin Wisp. We sample $5 \times 10^5$ points for each shape per epoch of training. We sample $10^5$ points uniformly in the domain $\Omega = \{\|\bm{x}\|_\infty \leq 1 \rvert \bm{x} \in \mathbb{R}^3\}$, $2 \times 10^5$ on the surface of the shape, and $2 \times 10^5$ near the surface using normal distribution with a variance of 0.015. We use the same input coordinates for the training of all INRs. We train the grid-based INRs for 10 epochs and the MLP-only INRs for 100 epochs. We use Adam optimizer with a learning rate of $1e-3$ ~\cite{adam}.

\noindent{\bf Compute Resources and Training Time} All experiments and speed measurements were conducted on an Ubuntu 22.04 LTS system equipped with an Intel i7-13700K CPU and an NVIDIA RTX 4090 GPU. The retrieval and conversion times reported in \cref{tab:sdf_retrieval}\,\ref{tab:cross_architecture}.\,\ref{tab:conversion_speed_matrix}.\,\ref{tab:category_chamfer_result}.\,were all measured on this system. We used Kaolin Wisp’s implementation unless otherwise specified. We identified a significant inefficiency in Kaolin Wisp's point sampling algorithm, which performs point batching in plain Python. We addressed this issue by performing batching in PyTorch, resulting in approximately a 10x speedup. Consequently, training an INR and converting it to other data types takes about 1 minute in total. Evaluating our methods requires 12 INRs per shape (across 4 architectures and 3 implicit functions), leading to approximately 8.3 GPU days for INR training in the ShapeNet10 experiments alone. We anticipate that further improvements in INR modeling could reduce training times for even larger-scale experiments.

\noindent{\bf INR Initialization.} For the MLP component of INR, we follow inr2vec's method and initialize the MLPs (of different INRs) with the same weights, as this has been proven essential for ensuring that the embedding from the MLP component is meaningful for shape retrieval~\cite{inr2vec}. However, we observe no such restriction for the initialization of the feature grids for different INRs, which is initialized with very small random values.

\subsection{INR Encoder Details}
\label{app:inr_encoder_arch}
\textbf{INR MLP Encoder.}
\label{app:inr_mlp_encoder_arch}
We use an MLP encoder to convert the weights of the INR MLP into an embedding. For encoding MLP-only INRs, we use an encoder that is itself an MLP. This MLP encoder consists of 4 linear layers, each followed by batch normalization and a ReLU activation~\cite{batchnorm}. The final layer is a max pooling layer that produces an embedding of length 1024. For encoding the MLP component of a grid-based INR, we reduce the hidden size of all layers by half, using hidden layers with sizes 256, 256, 512, and 512. This results in an INR MLP embedding of length 512.

\noindent{\bf INR Conv3D Encoder.}
\label{app:inr_conv3d_encoder_arch}
The Conv3D Encoder consists of five 3D convolution operations. Each convolution uses a kernel size of $(2, 2, 2)$ and a stride of $(2, 2, 2)$ to gradually reduce the spatial resolution. Each convolutional layer doubles the channel size and is followed by group normalization and a ReLU activation~\cite{groupnorm}. The final layer is a linear layer that maps the convolution output to an INR Grid Embedding of length 512. Combined with the INR MLP Embedding, the total INR Embedding length for the grid-based INR is 1024, which is the same as the embedding length for the MLP-only INR in inr2vec.

\noindent{\bf INR Encoder Training.}
We use the same input sampling process as in the INR training. Following the procedure in~\cite{inr2vec}, we use the AdamW optimizer with a learning rate of $1 \times 10^{-4}$ and a weight decay of $1 \times 10^{-2}$~\cite{adamw}. Note that our decoder MLP $f_\phi$ has the same architecture as in~\cite{inr2vec}. Only the encoders are necessary for generating the INR embeddings. The decoder is used solely during the training of the encoders, and is not needed when encoding INR Embedding during inference.

\subsection{INR Distillation}
\label{app:distillation}

Changes such as modifications to the feature grid dimensions or the number of hidden nodes in the MLP can cause dimension mismatches, making encoders trained for specific architectures unusable. However, since one might need different architecture configurations for trade-offs between speed and representation quality, or use architectures that may be developed in the future, we need a general solution that does not have strict requirements on the INR architecture.

To address this, we leverage the property of INRs designed to output distance or occupancy values given an input coordinate. For a source INR with an unknown architecture, we create an embedding for retrieval by using the source INR $f_s$ as an oracle. This involves generating pairs of input coordinates and output values from $f_s$ to train an INR $f_\theta$ with an architecture compatible with our encoders. We refer to this as the INR distillation technique, as illustrated in~ \cref{fig:inr_cross_arch}.

\begin{equation} \label{eq:remapping}
  f_\theta(\bm{x}; \bm{z}(\bm{x}, \mathcal{Z})) \approx f_s(\bm{x}), \,\forall \bm{x} \in \Omega.
\end{equation}

In general, there are no limitations on the source or target INR architectures or the type of output value (distance or occupancy). In~\cref{result:diff_arch}, we showed that distilling a source MLP-only INR to an INR with a feature grid can actually improve retrieval accuracy compared to using embeddings created from the MLP-only INR.

\subsection{Explicit Representation Training and Encoding}
\label{app:pointnext_viewgcn}
\noindent{\bf PointNeXt.} For training PointNeXt~\cite{ret_pc_pointnext}, we use point clouds containing 2048 points sampled from the surfaces of the INRs representing shapes in the training set. We follow the training procedure outlined in PointNeXt, using the PointNeXt-S variant. The training is supervised based on the shape class. After training, we remove the classification head and use the output from the PointNeXt backbone to create embeddings of length 512.

\noindent{\bf View-GCN.} For training View-GCN~\cite{ret_proj_view_gcn}, we use images rendered from the INRs representing the training shapes. We render 12 images at a resolution of 224 x 224 from virtual cameras positioned 3 units away from the object's center with a 0.65 elevation along a circular trajectory. We follow the training procedure specified in View-GCN, using shape class labels for supervision. After training, we remove the classification head and use the output of length 1536 as the embedding for shape retrieval.

\subsection{Relationship Between Different Implicit Functions}
\label{app:udf_sdf_relation}

At first glance, different implicit function fields may seem entirely distinct, even for similar shapes. For instance, the interior of a watertight shape is negative in an SDF representation but positive in a UDF representation. This raises the question of how shape encoders can map INRs with different implicit function fields to a shared embedding space.

We analyze whether standard neural networks can relate different implicit function values. For the same underlying shape, the UDF, SDF, and Occ values are related by the following equations:
\begin{equation}
\text{UDF} = \text{ReLU}(\text{SDF}) + \text{ReLU}(-\text{SDF})
\label{eq:udf_sdf_relationship}
\end{equation}
\begin{equation}
\text{Occ} = \text{Sign}(\text{SDF})
\label{eq:occ_sdf_relationship}
\end{equation}

~\cref{eq:udf_sdf_relationship} utilizes standard summation, multiplication, and ReLU activations. \cref{eq:occ_sdf_relationship} can also be calculated precisely using summation, multiplication, and ReLU activations, following these operations:
\begin{equation}
\begin{cases}
h_1 = \text{ReLU}(\text{SDF}) \\
h_2 = \text{ReLU}(-\text{SDF}) \\
h_3 = \text{ReLU}(h_1 - 1) \\
h_4 = \text{ReLU}(h_2 - 1)
\end{cases}
\end{equation}

\begin{equation}
\text{Occ} = h_1 - h_2 - h_3 + h_4
\end{equation}

\begin{table*}[h]
\centering
\begin{tabular}{c|c|c|c}
\hline
                             & {UDF}  &  {SDF}  & Occ  \\ \hline
Point Cloud       & {1.26} & {0.98} & 1.82 \\ 
Multi-View Images & {4.13} & {3.05} & 3.67 \\ \hline
\end{tabular}
\caption{Conversion Speed (seconds) from INR with Different Implicit Functions to Different Representations}
\label{tab:conversion_speed_matrix}

\centering
\begin{tabular}{c|c|c|c|c}
\hline

     \multicolumn{2}{c|}{}     & \multicolumn{3}{c}{Retrieval INR}                                       \\ \cline{3-5} 
      \multicolumn{2}{c|}{}                     & {UDF}            & {SDF}            & Occ            \\ \hline
\multirow{3}{*}{\rotatebox[origin=c]{90}{Query}} & UDF & {83.0/68.8/68.6} & {79.0/10.4/66.2}  & {80.4/8.8/66.6} \\  
                                                 & SDF & {81.8/11.4/67.8} & {84.6/70.2/70.0} & {82.4/10.4/67.4}  \\  
                                                 & Occ & {81.0/{\textcolor{white}0}9.2/67.2} & {82.6/10.4/68.0} & {83.0/69.4/78.6} \\ \hline
\multicolumn{2}{c|}{Average} & \multicolumn{3}{c}{82.0/29.9/67.8} \\
\hline
\multicolumn{2}{c|}{\textit{Legend}} & \multicolumn{3}{c}{iNGP / MLP-only / MLP-only + INRet Regularization} \\
\hline
\end{tabular}
\caption{Shape Retrieval Accuracy for Different Implicit Function INRs on ShapeNet10}
\label{tab:mlp_with_inret}

\end{table*}

Our INR Encoders do not learn these mappings directly, as they operate with learned INR features at a global scale. However,~\cref{eq:udf_sdf_relationship} demonstrates that learning similar or even identical representations from UDF, SDF, and Occ is theoretically possible with standard neural network operations.

We visualize the differences between implicit functions of a simplified shape in~\cref{fig:udfsdf}. Consider the cross-section of a watertight ball in~\cref{fig:udfsdf}(a) and (b). ~\cref{fig:udfsdf}(a) and (b) show the SDF and UDF field respectively, and these fields can be simply related by~\cref{eq:udf_sdf_relationship}. \cref{fig:udfsdf}(c) is the typical learned Occ field of the same ball, where the values near the surface is zero, but +1 or -1 elsewhere. Note that the exact Occ field should be a solid fill both inside and outside the surface, but INRs often have trouble learning these exact values perfectly, and often learn near-zero values around a small region of the surface, which we show here.

Lastly, we show the SDF field of the cross-section of a double layer cylinder (open top and bottom) in~\cref{fig:udfsdf}(d). Compare this with~\cref{fig:udfsdf}(a), which is the UDF cross-section of a single layer cylinder, the UDF and SDF fields are almost the same everywhere except for the region in between the two layers. Note that SDF can be not be calculated for the single layer cylinder due to the lack of a watertight surface. This similarities shows that for the same or very similar shape, the underlying implicit function fields are also very similar, making learning the same embedding for the different fields easier.

\begin{table*}
\centering
\begin{tabular}{c|c|c|c|c}
\hline

     \multicolumn{2}{c|}{}     & \multicolumn{3}{c}{Retrieval INR}                                       \\ \cline{3-5} 
      \multicolumn{2}{c|}{}                     & {UDF}            & {SDF}            & Occ            \\ \hline
\multirow{3}{*}{\rotatebox[origin=c]{90}{Query}} & UDF & {\textbf{83.0}/\underline{80.8}/79.4} & {79.0/81.2/78.8}  & {80.4/79.8/80.4} \\  
                                                 & SDF & {81.8/81.2/80.0} & {\underline{84.6}/\textbf{85.8}/83.6} & {82.4/82.4/82.6}  \\  
                                                 & Occ & {81.0/79.6/80.8} & {82.6/\textbf{82.8}/81.4} & {83.0/\underline{83.2}/\textbf{83.4}} \\ \hline
\multicolumn{2}{c|}{Average} & \multicolumn{3}{c}{\textbf{82.0}/\underline{81.9}/{81.2}} \\
\hline
\end{tabular}
\caption{Shape Retrieval Accuracy for iNGP INRs on ShapeNet10 with Different \textit{Unified Shape Decoder} Implicit Functions (UDF/SDF/Occ)}
\label{tab:decoder_choice}

\centering
\begin{tabular}{c|c|c|c|c}
\hline

     \multicolumn{2}{c|}{}     & \multicolumn{3}{c}{Retrieval INR}                                       \\ \cline{3-5} 
      \multicolumn{2}{c|}{}                     & {UDF}            & {SDF}            & Occ            \\ \hline
\multirow{3}{*}{\rotatebox[origin=c]{90}{Query}} & UDF & {83.0/82.6/82.0/82.8} & {79.0/80.2/78.4/80.4}  & {80.4/80.6/81.0/79.8} \\  
                                                 & SDF & {81.8/81.8/82.0/81.0} & {84.6/84.0/83.8/84.8} & {82.4/81.4/81.8/80.8}  \\  
                                                 & Occ & {81.0/81.2/80.8/80.6} & {82.6/82.0/83.0/82.6} & {83.0/82.6/83.2/82.8} \\ \hline
\multicolumn{2}{c|}{Average} & \multicolumn{3}{c}{\textbf{82.0}/\underline{81.8}/{81.8}/{81.7}} \\
\hline
\end{tabular}
\caption{Shape Retrieval Accuracy for iNGP INRs on ShapeNet10 with Different Explicit L2 Regularization Weights for \textit{UDF-SDF,\ UDF-Occ,\ SDF-Occ} (\textit{111/211/121/112})}
\label{tab:explicit_weight_choice}

\centering
\begin{tabular}{c|c|c|c|c}
\hline

     \multicolumn{2}{c|}{}     & \multicolumn{3}{c}{Retrieval INR}                                       \\ \cline{3-5} 
      \multicolumn{2}{c|}{}                     & {UDF}            & {SDF}            & Occ            \\ \hline
\multirow{3}{*}{\rotatebox[origin=c]{90}{Query}} & UDF & {83.0/82.5} & {79.0/81.0}  & {80.4/80.6} \\  
                                                 & SDF & {81.8/81.6} & {84.6/84.6} & {82.4/81.8}  \\  
                                                 & Occ & {81.0/81.4} & {82.6/79.0} & {83.0/79.6} \\ \hline
\multicolumn{2}{c|}{Average} & \multicolumn{3}{c}{\textbf{82.0}/\underline{81.3}} \\
\hline
\end{tabular}
\caption{Shape Retrieval Accuracy for iNGP INRs on ShapeNet10 with UDF \textit{Unified Common Decoder} L2/L1 Loss Choice}
\label{tab:UDF_common_decoder_l1l2}
\end{table*}

\section{Additional Results}
\label{app_sec:exp}

In this section of the appendix, we provide additional results and ablation studies for \name. App.~\ref{app_subsec:exp1} and \ref{app_subsec:exp3} provides additional results for retrieval accuracy evaluation on ShapeNet10 and Pix3D. App.~\ref{app_ssec:inret_on_mlp} demonstrates the effectiveness of INRet's regularizations on the retrieval of MLP-only INRs. App.~\ref{app:shape_decoder_implicit_choice} and \ref{app_ssec:norm_choice} examines the impact of the implicit function of the \textit{Unified Shape Decoder} on the final accuracy. App.~\ref{app_ssec:sum_vs_concat} examines the impact of summation \textit{vs.} concatenation of features from the spatial grids on the retrieval accuracy.

\subsection{INR Retrieval with Feature Grids}
\label{app_subsec:exp1}

In~\cref{tab:appendix_exp1_shapenet} and~\cref{tab:appendix_exp1_pix3d}, we provide additional results and metrics for the experiment listed in~\cref{exp:exp_grid}. We show the mean Average Precision (mAP@k) at different numbers of k following the method in~\cite{inr2vec}. We also report the precision, recall, and F1 score following the definition in ShapeNet~\cite{shrec16}. Note that for the ShapeNet10 dataset, since the number of models in each category is the same, the precision, recall and F1 score are the same, and thus we only report the F1 score. Our method achieves higher scores for almost all metrics across both ShapeNet10 and Pix3D datasets over inr2vec, PointNeXt and View-GCN. This demonstrates that our method is not only able to correctly retrieve the most similar shape, but also retrieves more shapes that belongs to the same category as the query shape as seen by the higher F1 score.

\subsection{INRs with Different Implicit Functions}
\label{app_subsec:exp3}

We show additional results for INR retrieval across different implicit functions on the Pix3D dataset in~\cref{tab:cross_retrieval_pix3d}. Similar to the results on the ShapeNet10 dataset, our method demonstrates higher accuracy for retrieval across INRs with different implicit functions than inr2vec, PointNeXt and View-GCN.

Normally, we exclude the INR representing the same shape from being retrieved when measuring the mAP, otherwise, the query embedding always have the highest cosine similarity with itself. In~\cref{tab:cross_implicit_retrieve_same_shape}, we show the accuracy of INR retrieval across different implicit functions by allowing retrieval of INR (with a different implicit function) representing the same shape. As seen in the table, there is around 10\% improvement in retrieval accuracy. This shows that in many cases, the retrieved shape is the same shape as the query shape, but just represented with a different implicit function.

In~\cref{tab:conversion_speed_matrix}, we show the conversion speed of converting different representations to point clouds and multi-view images. As required by View-GCN, 12 images are rendered, and as a result, it is more expensive than sampling a single point cloud. Conversion to point cloud or images is also more expensive for UDF compared to SDF due to the use of damped spherical tracing.

\subsection{Applying INRet Regularization to MLP-only INRs}
\label{app_ssec:inret_on_mlp}
In~\cref{tab:mlp_with_inret}, we demonstrate the retrieval accuracy when we apply INRet unified latent space regularizations (L2 + \textit{Unified Shape Decoder}) to MLP-only INRs. We also include the iNGP retrieval accuracy for comparison. As seen from the table, the retrieval accuracy of MLP-only INRs significantly increases when the unified latent space regularizations are applied. This shows that our regularization techniques apply to both INR with and without feature grids. However, for the MLP-only INRs, the final accuracy is still lower than if the iNGP INR with feature grid is used to create the INR embeddings.

\subsection{Choice of Unified Shape Decoder Implicit Function}
\label{app:shape_decoder_implicit_choice}
In this section, we evaluate the performance of \name with different Unified Shape Decoder implicit functions. In~\cref{ssec:inr_diff_imp}, we used the UDF as the implicit function for the \textit{Unified Shape Decoder}. In~\cref{tab:decoder_choice}, we show the retrieval accuracy when the unified decoder outputs different alternative implicit functions during training.

As seen in~\cref{tab:decoder_choice}, the average retrieval accuracy for different choices of common decoders is fairly close. The UDF common decoder had the highest accuracy of 82.0\% while the lowest, the Occ common decoder, is only 0.8\% behind in accuracy. However, we observe that if the INR's implicit function is the same as the common decoder's output, the retrieval accuracy tends to be higher. For example, for SDF to SDF retrieval, the highest retrieval accuracy of 85.8\% is achieved when the common decoder's output is also an SDF. The trend also applies to UDF to UDF retrieval and Occ to Occ retrieval. In addition, for retrieval across INRs with different implicit functions, the retrieval accuracy tends to be higher if the query or retrieval INR's implicit function is the same as the common decoder's output type. Despite these tendencies, our method is generally relatively robust to the choice of the common decoder's output.

\subsection{L2 Regularization Weight}
\label{app_ssec:l2_reg_weight}
In this section, we evaluate the performance when different weights are applied to the explicit L2 regularization. In \name, the explicit L2 regularization is simultaneously applied to 3 different pairs: UDF-SDF, UDF-Occ and SDF-Occ. In the main results presented in the paper, the weighting is the same for all pairs. In~\cref{tab:explicit_weight_choice}, we show the retrieval accuracy when the weights are different. For example, the \textit{211} weight means the UDF-SDF loss is multiplied by 2 before being added to the total loss, while the UDF-Occ and SDF-Occ are multiplied by 1.

From~\cref{tab:explicit_weight_choice}, we observe that our method is robust with respect to the specific choice of weight multipliers. For the INR encoder training, we used the Adam optimizer which has an adaptive learning rate on individual weights of the network, eliminating the need for careful fine-tuning on the weight multipliers~\cite{adam}. 

\subsection{Norm Choice for Unified Shape Decoder}
\label{app_ssec:norm_choice}

For a specific implicit function, our choice of norm simply follows that used in existing works. As explained in Appendix \ref{app_ssec:inr_training_detail}, we follow the method in~\cite{udf_neurips_2020} and use the L1 normalization for both UDF INR training and when the \textit{Unified Shape Decoder's} implicit function is UDF. In this section, we test whether using L2 normalization for the \textit{Unified Shape Decoder} (with UDF implicit function) instead of L1 has an impact on the accuracy. We present the results in~\cref{tab:UDF_common_decoder_l1l2}. From the table, we show that using L2 normalization decreases the retrieval accuracy slightly compared to using the L1 loss on average. We note that the retrieval accuracy of individual loss function to loss function pairs can fluctuate quite significantly. For example, the Occ-SDF retrieval accuracy dropped 3.6\% (from 82.6\% to 79.0\%). This is different from the result in~\cref{tab:explicit_weight_choice} where the weighting of the explicit regularization had minimal impact on the retrieval accuracy.

\subsection{Summation and Concatenation of Features}
\label{app_ssec:sum_vs_concat}
In the main results, the Conv3D encoder takes in the summation of features from NGLOD feature grid and the concatenation of features from iNGP feature grid. We do so because summation and concatenation of features are used in the original NGLOD INR and iNGP INR respectively.

Following~\cref{eq:decode_to_distance_value}, for NGLOD, the features from the multi-level octree feature grid are summed before fed into the MLP, i.e.
\begin{equation}
    \bm{z}(\bm{x}, \mathcal{Z}) = \sum_{l}^{L}(\bm{\psi}(\bm{x}; l, \mathcal{Z}))
\end{equation}
For iNGP, the features are concatenated instead, i.e.
\begin{equation}
    \bm{z}(\bm{x}, \mathcal{Z}) = [\bm{\psi}(\bm{x}; 0, \mathcal{Z}), \bm{\psi}(\bm{x}; 1, \mathcal{Z}), \ldots, \bm{\psi}(\bm{x}; L, \mathcal{Z})]
\end{equation}

For NGLOD, features stored in different levels of the octree capture varying levels of geometry detail. Therefore, using summation allows adding finer surface information (deeper level) to the coarser overall shape (upper level)~\cite{nglod}. For iNGP, the features stored in the hash grid inevitably suffer from hash collision. The authors argued that using features from all levels would allow the MLP to mitigate the effect of hash collision dynamically~\cite{instantngp}. Using the octree feature grid, NGLOD does not suffer from the hash collision issue.

Following the experiment setting listed in~\cref{exp:exp_grid}, we test the retrieval accuracy when we use features in a way different from how it was used in the original INR architecture.

\begin{table*}[ht]
\centering
\begin{tabular}{c|cc|cc}
\hline
INR Arch.     & \multicolumn{2}{c|}{NGLOD}                              & \multicolumn{2}{c}{iNGP}                                \\ \hline
Feature Comb. & \multicolumn{1}{c|}{Sum (Original)} & Concat (Modified) & \multicolumn{1}{c|}{Concat (Original)} & Sum (Modified) \\ \hline
mAP @ 1       & \multicolumn{1}{c|}{\textbf{82.6}}  & 67.8              & \multicolumn{1}{c|}{\textbf{84.2}}     & 30.4           \\ \hline
\end{tabular}
\caption{Shape Retrieval Accuracy on ShapeNet10 when Features are Summed or Concatenated from the Feature Grids}
\label{tab:sum_vs_concat}
\end{table*}

As shown in~\cref{tab:sum_vs_concat}, both methods experienced a significant drop in retrieval accuracy if the features were not used in a way consistent with the original INR. For NGLOD, the concatenation of features leads to an accuracy drop of 14.8\%. We note that the concatenation of features in this case actually means more features being passed to the Conv3D encoder for INR embedding creation. However, since the finer level features were never used alone in the original NGLOD INR training, we hypothesize the Conv3D encoder may be overfitting to these finer level features that might be noisy when used standalone. For iNGP, the retrieval accuracy is dropped by 53.8\% since the summation of features leads to a significant loss of information.

\subsection{Additional Retrieval Visualization}
\begin{figure}[h]
\centering
    \includegraphics[width=\linewidth]{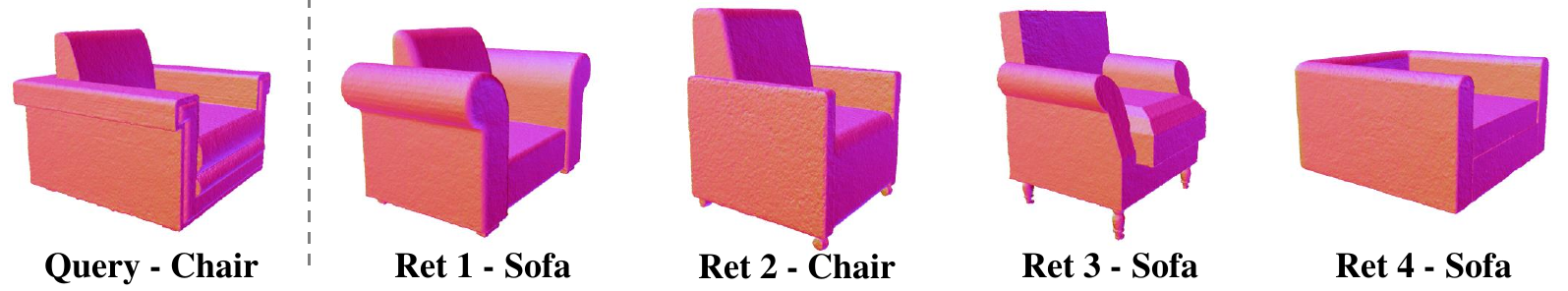}
    \caption{Chair Retrieval Incorrect Classes}
    \label{fig:chair_retrieval}
\end{figure}

We show retrievals that failed to retrieve from the same category in~\cref{fig:chair_retrieval}. As seen in the figure, given a query chair, the retrieved examples can be from other categories albeit resembling some semantic similarities with the query itself.

\section{the Impact of Reconstruction Quality on Retrieval Accuracy}
\subsection{Reconstruction Quality}
\label{app_ssec:recon_quality}

In this section, we provide additional details on the quality of reconstruction of the trained INRs with respect to the original mesh. For UDF INRs, we measure the Chamfer Distance (C.D.) at 130,172 points, following the same sampling method used in~\cite{nglod}. However, instead of regular spherical tracing, we apply damped spherical tracing similar to~\cite{udf_neurips_2020}. For SDF and Occ INRs, we use vanilla spherical tracing without damping, and we also measure generalized Intersection over Union (gIoU) which calculates the intersection of the inside of two watertight surfaces with respect to their union. We do not measure gIoU for UDF INRs as there is no notion of inside and outside.

\begin{table*}[ht]
\centering
\small
\begin{tabular}{c|ccc|ccc|ccc}
\hline
INR Arch.           & \multicolumn{3}{c|}{NGLOD}                                                   & \multicolumn{3}{c|}{iNGP}                                                                            & \multicolumn{3}{c}{MLP}                                                             \\ \hline
Implicit Func.      & \multicolumn{1}{c|}{SDF}        & \multicolumn{1}{c|}{UDF}          & Occ    & \multicolumn{1}{c|}{SDF}           & \multicolumn{1}{c|}{UDF}             & \multicolumn{1}{l|}{Occ} & \multicolumn{1}{c|}{SDF}    & \multicolumn{1}{c|}{UDF}    & \multicolumn{1}{l}{Occ} \\ \hline
C.D. ShapeNet     & \multicolumn{1}{c|}{0.0168}     & \multicolumn{1}{c|}{\underline{0.0122}} & 0.0210 & \multicolumn{1}{c|}{0.0147}        & \multicolumn{1}{c|}{\textbf{0.0119}} & 0.0223                   & \multicolumn{1}{c|}{0.0354} & \multicolumn{1}{c|}{0.0344} & 0.0389                  \\
C.D. Pix3D          & \multicolumn{1}{c|}{0.0183}     & \multicolumn{1}{c|}{\underline{0.0125}} & 0.0213 & \multicolumn{1}{c|}{0.0146}        & \multicolumn{1}{c|}{\textbf{0.0120}} & 0.0241                   & \multicolumn{1}{c|}{0.0367} & \multicolumn{1}{c|}{0.0351} & 0.0392                  \\
gIoU ShapeNet & \multicolumn{1}{c|}{\underline{84.2}} & \multicolumn{1}{c|}{NA}           & 81.4   & \multicolumn{1}{c|}{\textbf{86.2}} & \multicolumn{1}{c|}{NA}              & 82.1                     & \multicolumn{1}{c|}{77.3}   & \multicolumn{1}{c|}{NA}     & 75.2                    \\
gIoU Pix3D      & \multicolumn{1}{c|}{\underline{85.5}} & \multicolumn{1}{c|}{NA}           & 82.2   & \multicolumn{1}{c|}{\textbf{86.5}} & \multicolumn{1}{c|}{NA}              & 82.3                     & \multicolumn{1}{c|}{77.5}   & \multicolumn{1}{c|}{NA}     & 74.9                    \\ \hline
\end{tabular}
\caption{Shape Reconstruction Quality of different INRs on ShapeNet and Pix3D}
\label{tab:reconstruction_quality}
\end{table*}

As seen in~\cref{tab:reconstruction_quality}, both the NGLOD and iNGP achieve higher reconstruction quality than the MLP INRs. These INRs with feature grids are not only superior at representing shapes with higher fidelity but also lead to higher retrieval accuracy.

\subsection{Reconstruction Quality and Retrieval Accuracy}
\label{app_ssec:quality_and_accuracy}
Since INR with feature grids have both higher reconstruction quality and higher retrieval accuracy, one may wonder if these are correlated. We perform another experiment, where the iNGP is only trained for only 2 epochs, leading to reconstruction quality lower than the MLP-only INR. As seen in~\cref{tab:ingp_2epoch}, the retrieval accuracy for iNGP significantly drops when the INRs are undertrained. However, retrieval with iNGP @ 2 epochs still has 5.4\% higher accuracy compared to retrieval with MLP-only INR. The MLP-only INR lacks the features stored spatially in the feature grid, which is very useful for improving retrieval accuracy.

\begin{figure*}[h]
\centering
    \includegraphics[width=\linewidth]{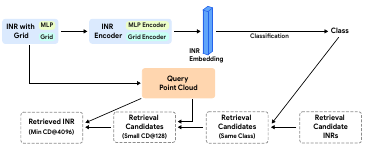}
    \caption{Hierarchical Sampling Retrieval Method}
    \label{fig:rebuttal_chamfer}
\end{figure*}

\begin{table*}[ht]
\centering
\begin{tabular}{c|cc|c}
\hline
Method     & \multicolumn{2}{c|}{Ours}                                                               & inr2vec \\ \hline
Input Type & \multicolumn{1}{c|}{iNGP}                           & iNGP @ 2 Epoch                    & MLP INR \\ \hline
mAP @ 1    & \multicolumn{1}{c|}{\textbf{84.2}} & \underline{78.8} & 73.4    \\
C.D.       & \multicolumn{1}{c|}{0.0168}                         & 0.0371                            & 0.0354  \\ \hline
\end{tabular}
\caption{Shape Retrieval Accuracy and Reconstruction Quality Comparison for Different INR Architectures (SDF) on ShapeNet10}
\label{tab:ingp_2epoch}

\centering
\begin{tabular}{c|c|ccc}
\hline
\multicolumn{2}{c|}{\multirow{2}{*}{}}                & \multicolumn{3}{c}{Retrieval INR}                                                                                   \\ \cline{3-5} 
\multicolumn{2}{c|}{}                           & \multicolumn{1}{c|}{UDF}                    & \multicolumn{1}{c|}{SDF}                     & Occ                     \\ \hline
\multirow{3}{*}{\rotatebox[origin=c]{90}{Query}} & UDF & 
\multicolumn{1}{c|}{80.2($-$0.2) / 83.0($+$0.0)} & \multicolumn{1}{c|}{91.2($+$0.2) / 88.4($-$0.2)}   & 86.6($+$0.0) / 87.4($-$0.4)   \\ \cline{2-5}
& SDF & 
\multicolumn{1}{c|}{92.0($+$0.0) / 93.2($-$0.2)} & \multicolumn{1}{c|}{83.4 / 84.6}               & 90.2 / 92.8 \\ \cline{2-5}
& Occ & 
\multicolumn{1}{c|}{85.6($+$0.0) / 89.4($+$0.0)}  & \multicolumn{1}{c|}{87.8 / 92.6} & 76.8 / 83.0               \\ \hline
\end{tabular}
\caption{Shape Retrieval Accuracy for Different Implicit Function INRs on ShapeNet10 (Accuracy Change when UDF INRs are trained using point clouds instead of meshes)}
\label{tab:cross_retrieval_pc_as_udf_input}
\end{table*}

\section{Retrieval of INRs trained using Different Source Data}
\label{app_sec:udf_pc}

In Section \ref{ssec:inr_diff_imp}, we demonstrated the retrieval accuracy across different INR implicit functions. These implicit functions are trained using the same source information (meshes). In~\cref{tab:cross_retrieval_pc_as_udf_input}, we show another case where the UDF INRs are trained using point clouds sampled from the meshes instead of using the meshes directly~\cite{udf_neurips_2020}. As seen in~\cref{tab:cross_retrieval_pc_as_udf_input}, the retrieval accuracy is very similar regardless of the type of the source training data, showing that \name can enable the retrieval of INRs when the INRs are trained with different inputs.

\section{Category-Chamfer Metric}
\label{app:category_chamfer}

\subsection{Retrieval Accuracy by Category and Chamfer Distance}
Shape retrieval performance is traditionally evaluated based on whether the retrieved shape has the same category as the query shape~\cite{shrec16}. While this metric can evaluate the quality of retrieval based on overall shape semantics, it largely ignores similarities or differences between individual shape instances. To mitigate the shortcomings of existing metrics, we propose the Category-Chamfer metric, which evaluates whether the retrieved shape shares the same category as the query shape, and also has the least Chamfer Distance with respect to the query shape for all shapes in the category.

We choose Chamfer Distance as it can measure the distance between almost all 3D representations. Chamfer Distance is a metric that calculates similarity between two point clouds. Unlike certain metrics such as generalized Intersection over Union (gIoU) that require watertight surfaces with a clear definition of inside and outside, Chamfer Distance only requires a point cloud which can be easily converted from other 3D representations including meshes, voxels, and INRs.

The accuracy $A_C$ based on category information only is
\begin{equation}
A_C = \frac{\sum_{q \in Q} \delta(C(q), C(R(q)))}{|Q|}
\end{equation}
where $Q$ is the query set, $C$ and $R$ denote the category and retrieval function respectively, the Kronecker delta $\delta(\cdot, \cdot)$ evaluates to $1$ if $C(q)$ and $C(R(q))$ are the same and $0$ otherwise. The accuracy is normalized by the total length $|Q|$ of the query set.

The accuracy $A_{CC}$ based on category and Chamfer Distance is

\begin{multline}
A_{CC} = \frac{\sum_{q \in Q} \left[ \delta(C(q), C(R(q))) \times \delta\left(s', C(R(q))\right) \right]}{|Q|} \\
\text{where } s' = \operatorname*{argmin}_{s \in S} d_{CD}(q, s)
\end{multline}

where $d_{CD}$ denotes the Chamfer Distance, $S$ denotes the set of all candidates for retrieval.

Category-Chamfer is a more challenging metric compared to category-only metric, in our experiments, we find that we can leverage the Chamfer Distance between the the INR instances to achieve a high accuracy for this metric.

\subsection{Category-Chamfer Retrieval Accuracy by Embedding Cosine Similarity}
Compared with the category-only accuracy, achieving high accuracy as measured by the Category-Chamfer metric is more challenging. By simply comparing the cosine similarity between embeddings, neither INRet or existing methods such as PointNeXt perform well for this new metric. We exclude View-GCN from this evaluation since it may not require an actual 3D model to perform the retrieval and thus may not be able to calculate Chamfer Distance given its input. Following the procedure in~\cref{result:diff_arch}, we evaluate the Category-Chamfer accuracy.

\begin{table*}[ht]
\centering
\begin{tabular}{c|cc|c}
\hline
Method     & \multicolumn{2}{c|}{Ours}                                                               & PointNeXt \\ \hline
Input Type & \multicolumn{1}{c|}{NGLOD}                           & iNGP                    & Point Cloud \\ \hline
$A_C$    & \multicolumn{1}{c|}{\underline{82.6}}                     & \textbf{84.2} & 71.2    \\
$A_{CC}$       & \multicolumn{1}{c|}{21.2}                        & \underline{23.2}                            & \textbf{28.4}  \\ \hline
\end{tabular}
\caption{Retrieval Accuracy (Category, Category-Chamfer) on ShapeNet10}
\label{tab:category_chamfer_accuracy_preliminary}

\centering
\begin{tabular}{c|clcl|cl}
\hline
Method                            & \multicolumn{4}{c|}{Ours}                                                                    & \multicolumn{2}{c}{PointNeXt}      \\ \hline
Input Type                        & \multicolumn{2}{c|}{NGLOD}                              & \multicolumn{2}{c|}{iNGP}          & \multicolumn{2}{c}{Point Cloud}    \\ \hline
$A_{CC}$                          & \multicolumn{2}{c|}{81.8}                               & \multicolumn{2}{c|}{82.4}          & \multicolumn{2}{c}{72.6}           \\ \hline
Ret. Time (Naive)                 & \multicolumn{2}{c|}{65.06}                              & \multicolumn{2}{c|}{65.06}         & \multicolumn{2}{c}{65.06}          \\ \hline
Ret. Time (Hier. Samp.) Total     & \multicolumn{2}{c|}{36.19}                              & \multicolumn{2}{c|}{35.46}         & \multicolumn{2}{c}{35.78}          \\
Ret. Time (Hier. Samp.) CD@128/4096 & \multicolumn{1}{c|}{25.05} & \multicolumn{1}{l|}{11.14} & \multicolumn{1}{c|}{25.05} & 10.41 & \multicolumn{1}{c|}{25.05} & 10.73 \\ \hline
\end{tabular}
\caption{Category-Chamfer Retrieval Accuracy and Retrieval Time on ShapeNet10}
\label{tab:category_chamfer_result}

\end{table*}

We calculate the ground truth Chamfer Distance at 131072 points following the same sampling method from~\cite{nglod}. From~\cref{tab:category_chamfer_accuracy_preliminary}, we observe that the Category-Chamfer accuracy for all methods is very low. The highest accuracy is achieved by PointNext at 28.4\%, far below its category-only accuracy of 71.2\%. In the next section, we provide a solution for increasing the Category-Chamfer retrieval accuracy while avoiding significant runtime overhead.

\subsection{Hierarchical Sampling}
Deep learning-based shape retrieval methods usually involve calculating an embedding for the input shape, and retrieval is done by comparing the cosine similarity between the embeddings. However, as seen in~\cref{tab:category_chamfer_accuracy_preliminary}, these methods do not perform well on the Category-Chamfer metric. Unlike cosine similarity which can be easily computed in a batched manner, Chamfer Distance requires comparison between individual point clouds. A naive solution is to calculate the Chamfer Distance with all other shapes within the same category. However, such a naive method would require extensive computation, scaling linearly with the size of the dataset for retrieval.

To this end, we propose a Hierarchical Sampling approach, visualized in~\cref{fig:rebuttal_chamfer}. We found that the Chamfer Distance at a small number of points (128) is an effective proxy for the Chamfer Distance at a large number of points (4096). Although we calculate the groundtruth Chamfer Distance at 131072 points following typical values used for evaluation of 3D shape reconstruction quality~\cite{nglod}, we found that in terms of ranking of shape by Chamfer Distance, 4096 points is sufficient. For INRet, we use the frozen INR Embeddings to train an MLP for classification, following the same settings as~\cite{inr2vec}. We use E-Stitchup to augment the input with interpolations of INR embeddings from the same class~\cite{estitchup}. For PointNeXt, we use the trained PointNeXt to do the classification.

We present the result of the retrieval in~\cref{tab:category_chamfer_result}. For naive retrieval, we directly sample points and calculate the Chamfer Distance at 4096 points between the query INR and all candidate INRs. For Hierarchical Sampling retrieval, we first sample points and calculate the Chamfer Distance at 128 points between the query INR and all candidate INRs. We further calculate the Chamfer Distance at 4096 points for all INRs with a small Chamfer Distance at 128 points. We define small by the INR having Chamfer Distance within 3 times of the smallest Chamfer Distance between query INR and all candidate INRs. This is a very generous bound and ensures a 100\% recall on our dataset. The accuracy is effectively only limited by the classification accuracy.

As shown in~\cref{tab:category_chamfer_result}, using Hierarchical Sampling significantly reduces the time (on average 1.8X) required for calculating the Chamfer Distance between different INRs. The speed-up for all methods is very similar as the point sampling and Chamfer Distance calculation time dominates the runtime. This leaves the difference in time for classification between the methods negligible. Using NGLOD as an example, the naive retrieval method involves point sampling and Chamfer Distance calculation (4096 points) for 49 INRs which costs 65.06 seconds, and an additional 0.04 seconds for classification. Using the hierarchical method, the distance point sampling and Chamfer Distance calculation are first done for 128 points (25.05 seconds + 0.04 seconds for classification), and around 17.1\% of the INRs need to be further evaluated at 4096 points, resulting in a runtime of 11.14 seconds. We expect this speedup to scale further as more data is presented as the retrieval candidate. Despite the speed up, this process is still relatively slow compared to the category-only retrieval which typically only requires cosine similarity comparison. We leave potential methods that would allow fast and accurate Category-Chamfer retrieval as future work.

\end{document}